\documentclass[letterpaper, 10 pt, final, conference, onside, twocolumn, nofonttune]{ieeeconf}  

\IEEEoverridecommandlockouts                              

\usepackage{graphics} 
\usepackage{epsfig} 
\usepackage{mathptmx} 
\usepackage{times} 
\usepackage{amsmath} 
\usepackage{leftidx}
\usepackage{amssymb}  
\usepackage{bm}  
\usepackage{multirow}
\usepackage{tikz}
\usepackage{subcaption}
\usepackage{listings,comment}
\usepackage[normalem]{ulem}
\usetikzlibrary{arrows,positioning,chains,fit,shapes} 
\tikzset{
    >=stealth',
    punkt/.style={
           rectangle,
           rounded corners,
           draw=black, very thick,
           text width=12em,
           minimum height=2em,
           text centered},
    pil/.style={
           ->,
           thick,
           shorten <=2pt,
           shorten >=2pt,},
	cross/.style={
    	cross out, 
        draw=black, 
        minimum size=2*(#1-\pgflinewidth), 
        inner sep=0pt, 
        outer sep=0pt},
	cross/.default={1pt}
}

\usepackage{amsthm}

\theoremstyle{plain}

\newtheorem{lemma}{Lemma}

\newtheorem{theorem}{Theorem}

\theoremstyle{definition}
\newtheorem{definitionx}{Definition} 
 \newenvironment{definition}
   {\pushQED{\qed}\definitionx}
   {\popQED\enddefinitionx}
\newtheorem{problemx}{Problem}
\newenvironment{problem}
  {\pushQED{\qed}\problemx}
  {\popQED\endproblemx}
\newtheorem{examplex}{Example}
\newenvironment{example}
  {\pushQED{\qed}\examplex}
  {\popQED\endexamplex}
 
\theoremstyle{remark}
\newtheorem{remark}{Remark}

\usepackage{algorithm,algorithmic}

\usepackage{xifthen}
\usepackage[symbols,toc]{glossaries-extra}

\makenoidxglossaries

\setabbreviationstyle[acronym]{long-short}
\preto\chapter{\glsresetall}

\newacronym{cegis}{CEGIS}{counterexample-guided inductive synthesis}
\newacronym{csp}{CSP}{constraint satisfiability problem}
\newacronym{smt}{SMT}{satisfiability modulo theories}
\newacronym{lp}{LP}{linear programming}
\newacronym{milp}{MILP}{mixed-integer linear programming}
\newacronym{ips}{IPS}{intelligent physical system}
\newacronym{ltl}{LTL}{linear temporal logic}
\newacronym{stl}{STL}{signal temporal logic}
\newacronym{mpc}{MPC}{model predictive control}

\glsxtrnewsymbol 
[description={logical tautology}]
{true}
{\ensuremath{\top}}

\glsxtrnewsymbol 
[description={logical contradiction}]
{false}
{\ensuremath{\perp}}

\glsxtrnewsymbol 
[description={logical conjunction}]
{and}
{\ensuremath{\wedge}}

\glsxtrnewsymbol 
[description={logical disjunction}]
{or}
{\ensuremath{\vee}}

\glsxtrnewsymbol 
[description={logical negation}]
{neg}
{\ensuremath{\neg}}

\glsxtrnewsymbol 
[description={logical conditional}]
{implies}
{\ensuremath{\rightarrow}}

\glsxtrnewsymbol 
[description={logical biconditional}]
{iff}
{\ensuremath{\leftrightarrow}}

\glsxtrnewsymbol
[description={the set containing no elements}]
{emptyset}
{\ensuremath{\emptyset}}

\glsxtrnewsymbol
[description={a propositional variable}]
{propvar}
{\ensuremath{p}}

\glsxtrnewsymbol 
[description={a predicate}]
{pred}
{\ensuremath{\pi}}

\glsxtrnewsymbol 
[description={a set of predicates}]
{predset}
{\ensuremath{\Pi}}

\glsxtrnewsymbol 
[description={assertion}]
{assert}
{\ensuremath{\leftarrow}}
\DeclareMathAlphabet\mathbfcal{OMS}{cmsy}{b}{n}

\glsxtrnewsymbol 
[description={the vector of continuous state variables}]
{statec}
{\ensuremath{\boldsymbol{x}}}

\glsxtrnewsymbol 
[description={the vector of initial continuous state}]
{statecini}
{\ensuremath{\bar{\boldsymbol{x}}}}

\glsxtrnewsymbol 
[description={length of the vector of continuous state variables}]
{statecnb}
{\ensuremath{n}}

\glsxtrnewsymbol 
[description={a vector of continuous input variables}]
{inputc}
{\ensuremath{\boldsymbol{u}}}

\glsxtrnewsymbol 
[description={length of the vector of continuous input variables}]
{inputcnb}
{\ensuremath{m}}

\glsxtrnewsymbol 
[description={the sampling period}]
{ts}
{\ensuremath{T_s}}

\glsxtrnewsymbol 
[description={a sequence of states and input variables evaluations}]
{run}
{\ensuremath{\boldsymbol{\xi}}}

\glsxtrnewsymbol 
[description={a sequence of states and input variables evaluations for a simulation model}]
{run_coarse}
{\ensuremath{\hat{\boldsymbol{\xi}}}}

\glsxtrnewsymbol 
[description={an \gls*{stl} formula}]
{stlformula}
{\ensuremath{\varphi}}

\glsxtrnewsymbol 
[description={a linear affine function}]
{stlfunc}
{\ensuremath{\mu}}

\glsxtrnewsymbol 
[description={a binary logic operator which states that an interpretation satisfies a formula}]
{sat}
{\ensuremath{\vDash}}

\glsxtrnewsymbol 
[description={the temporal logic operator always}]
{always}
{\ensuremath{\square}}

\glsxtrnewsymbol 
[description={the temporal logic operator eventually}]
{eventually}
{\ensuremath{\Diamond}}

\glsxtrnewsymbol 
[description={the temporal logic operator until}]
{until}
{\ensuremath{\boldsymbol{U}}}

\glsxtrnewsymbol 
[description={a real-valued function which associates a scalar value with the quality of satisfaction of an \gls*{stl} formula}]
{robfunc}
{\ensuremath{\rho}}

\glsxtrnewsymbol 
[description={a finite set of variables}]
{cspvarset}
{\ensuremath{V}}

\glsxtrnewsymbol 
[description={a variable}]
{cspvar}
{\ensuremath{v}}

\glsxtrnewsymbol 
[description={a finite set of domains of values}]
{cspdomset}
{\ensuremath{D}}

\glsxtrnewsymbol 
[description={a domainsof values}]
{cspdom}
{\ensuremath{\mathcal{D}}}

\glsxtrnewsymbol 
[description={a finite set of constraints}]
{cspconsset}
{\ensuremath{C}}

\glsxtrnewsymbol 
[description={a constraint}]
{cspcons}
{\ensuremath{c}}

\glsxtrnewsymbol 
[description={a real-valued function which associates a scalar value with the quality of satisfaction of an \gls*{stl} formula}]
{cspprob}
{\ensuremath{\langle \gls*{cspvarset}, \gls*{cspdomset}, \gls*{cspconsset} \rangle}}

\glsxtrnewsymbol 
[description={tolerance for dynamical feasibility}]
{tolfeas}
{\ensuremath{\delta}}

\glsxtrnewsymbol 
[description={a time indexed sequence of feedback controllers}]
{fctrler}
{\ensuremath{\boldsymbol{F}}}

\glsxtrnewsymbol 
[description={a time indexed sequence of convex constraints}]
{polyseq}
{\ensuremath{\boldsymbol{\mathbfcal{P}}}}

\glsxtrnewsymbol 
[description={a convex constraint}]
{poly}
{\ensuremath{\mathcal{P}}}

\glsxtrnewsymbol 
[description={an abstraction function}]
{abs}
{\ensuremath{\alpha}}

\glsxtrnewsymbol 
[description={a transition system}]
{tsys}
{\ensuremath{TS}}

\glsxtrnewsymbol 
[description={a time length}]
{tlen}
{\ensuremath{\ell}}

\glsxtrnewsymbol 
[description={a vector of time lengths}]
{tlenvec}
{\ensuremath{\boldsymbol{\ell}}}

\glsxtrnewsymbol 
[description={the prefix index}]
{preidx}
{\ensuremath{N}}

\glsxtrnewsymbol 
[description={the discrete plan robust measure function}]
{dplanrobfunc}
{\ensuremath{\nu}}

\glsxtrnewsymbol 
[description={the conservative maximum time length required to decide the satisfiability of a \gls*{stl} formula}]
{kmax}
{\ensuremath{K_{\max}^{\gls*{stlformula}}}}

\glsxtrnewsymbol 
[description={a slack variable}]
{slack}
{\ensuremath{s}}

\newacronym{idstl}{idSTL}{iterative deepening Signal Temporal Logic}

\makeatletter
\newcommand{\pushright}[1]{\ifmeasuring@#1\else\omit\hfill$\displaystyle#1$\fi\ignorespaces}
\newcommand{\pushleft}[1]{\ifmeasuring@#1\else\omit$\displaystyle#1$\hfill\fi\ignorespaces}
\makeatother

\DeclareMathOperator*{\argmax}{arg\,max}

\newcommand{\idstlpred}[1][]{\ensuremath{\gls*{pred}^{\ifthenelse{\isempty{#1}}{\gls*{stlfunc}}{#1}}}}
\newcommand{\idstlalways}[2]{\ensuremath{\gls*{always}_{#1} #2}}
\newcommand{\idstlevent}[2]{\ensuremath{\gls*{eventually}_{#1} #2}}
\newcommand{\idstluntil}[3]{\ensuremath{#1 \gls*{until}_{#2} #3}}
\newcommand{\idstlrobfunc}[2][]{\ensuremath{\gls*{robfunc}_{#1}^{#2}}}
\newcommand{\idstlinput}{\ensuremath{(\gls*{stlformula},A,B,\gls*{statecini},\gls*{ts},\gls*{tolfeas})}}
\newcommand{\idstloutput}{\ensuremath{\langle \text{Status}, \gls*{fctrler}, \gls*{run}, \gls*{polyseq}  \rangle}}
\newcommand{\idstldplaninput}{\ensuremath{(\gls*{stlformula},\gls*{statecini},\gls*{ts},\gls*{tolfeas})}}
\newcommand{\idstldplanoutput}{\ensuremath{\langle \text{Status},\gls*{polyseq},\gls*{run_coarse},L \rangle}}
\newcommand{\idstlfeasinput}{\ensuremath{(\gls*{stlformula},\gls*{polyseq},L,A,B,\gls*{statecini},\gls*{tolfeas})}}
\newcommand{\idstlfeasoutput}{\ensuremath{\langle \text{isFeasible},\gls*{run},\gls*{tlenvec},\gls*{polyseq}_{cex} \rangle}}
\newcommand{\idstlrobinput}{\ensuremath{(\gls*{stlformula},\gls*{polyseq},\gls*{tlenvec},L,Q_f,Q,R,A,B,\gls*{statecini},\gls*{tolfeas})}}
\newcommand{\idstlroboutput}{\ensuremath{\langle \gls*{fctrler}, \gls*{run},\gls*{tlenvec} \rangle}}
\newcommand{\encforma}[1]{\left|\left[#1\right]\right|}
\newcommand{\encformb}[1]{\left|\left\langle#1\right\rangle\right|}
\newcommand{\encformc}[1]{\left|\left(#1\right)\right|}

\makeatletter
\newcommand\fs@betterruled{%
  \def\@fs@cfont{\bfseries}\let\@fs@capt\floatc@ruled
  \def\@fs@pre{\vspace*{5pt}\hrule height.8pt depth0pt \kern2pt}%
  \def\@fs@post{\kern2pt\hrule\relax}%
  \def\@fs@mid{\kern2pt\hrule\kern2pt}%
  \let\@fs@iftopcapt\iftrue}
\floatstyle{betterruled}
\restylefloat{algorithm}
\makeatother

\makeatletter
\let\NAT@parse\undefined
\makeatother
\usepackage{hyperref}  

\title{ \LARGE \bf
Scalable Symbolic Control from Signal Temporal Logic Specifications
}

\author{Rafael Rodrigues da Silva$^{1,2}$ Vince Kurtz$^1$, and Hai Lin$^{1}$
	\thanks{$^{1}$ All authors are with Department of Electrical Engineering, University of Notre Dame, Notre Dame, IN 46556, USA.
		{\tt\small (rrodri17@nd.edu;~vkurtz@nd.edu;~hlin1@nd.edu)}}
\thanks{$^{2}$ The first author would like to appreciate the scholarship support by CAPES/BR, 99999.013242/2013-0}
}

\begin{document}

\maketitle
\thispagestyle{empty}
\pagestyle{empty}

\begin{abstract}

Many safety-critical systems must achieve high-level task specifications with guaranteed safety and correctness. Much recent progress towards this goal has been made through controller synthesis from \gls*{stl} specifications. Existing approaches, however, either consider some a priori discretization of the state-space, deal only with a convex fragment of \gls*{stl}, or are not provably complete. We propose a scalable, provably complete algorithm that directly synthesizes continuous trajectories to satisfy non-convex \gls*{stl} specifications. We separate discrete task planning and continuous motion planning on the fly and harness highly efficient \gls*{smt} and \gls*{lp} solvers to find dynamically feasible trajectories for high dimensional systems that satisfies non-convex \gls*{stl} specifications. The proposed design algorithms are proved sound and complete, and simulation results demonstrate the scalability of our approach.   
\end{abstract}

\section{INTRODUCTION}

\subsection{Motivation}

Autonomous \glspl*{ips} must be capable of interpreting high-level, possibly vague, tasks specifications, planning to achieve these specifications, and executing actions appropriate to a particular context in which the system is operating. \textit{Symbolic control} proposes to fulfill this need by automatically designing feedback controllers that result in the satisfaction of formal logic specifications. Temporal logics such as \gls*{stl} can express a wide variety of tasks for \glspl*{ips} \cite{baier2008principles}. Furthermore \gls*{stl} formulas are close to natural language and can even be interpreted from oral commands \cite{finucane2010ltlmop}. However, today's large-scale \glspl*{ips} present unprecedented challenges for symbolic control techniques, and existing \gls*{stl}-based symbolic control cannot solve many real-world problems.

A significant difficulty stems from the need to combine logical constraints (from task specifications) with continuous motion restrictions (from physical system dynamics). This integration is even more challenging for complex, high-dimensional systems which must execute tasks with non-convex logical constraints. For \gls*{stl} formulas based on convex predicates, operations like logical conjunction and negation preserve convexity in the overall formula. Disjuction operators and nested existential quantifications, however, render such an \gls*{stl} formula non-convex. For this reason, many symbolic control methods (e.g. \cite{raman2015reactive,lindemann2019control}) consider only a convex fragment of \gls*{stl}, though this restriction limits the expressiveness of the formal specification. 

Early efforts in symbolic control relied on discrete abstractions of continuous dynamical systems, and much work has focused on how to obtain an equivalent discrete and finite quotient transition system. Given an equivalent transition system, logical constraints can be handled with efficient search techniques in the discrete space. Finding such discrete abstractions is difficult in higher-dimensional spaces, however, and these approaches are usually limited to low-dimensional systems with no more than five continuous state variables \cite{rungger2013specification}. 

Recently, a growing body of work has focused on the synthesis of continuous trajectories from high-level logical specifications more directly. 
When considering only a convex fragment of \gls*{stl}, the problem can be efficiently encoded to a \gls*{lp} problem \cite{raman2014model}. Furthermore, the satisfaction of an \gls*{stl} formula can be measured using robust semantics, which allows efficient reactive control synthesis from control barrier functions \cite{lindemann2019control} and prescribed performance control \cite{lindemann2017prescribed}. However, the convex fragment of \gls*{stl} cannot describe many real-world applications of \gls*{ips}. For instance, a quadrotor performing an automated tasks needs to return to a charging station in reasonable time infinitely often: this desirable property that requires nested existential quantifications and thus cannot be described by a convex fragment of \gls*{stl}.  

\subsection{Our Idea}

We propose an efficient method of synthesizing dynamically-feasible trajectories that satisfy high-level \gls*{stl} specifications. Our approach does not require a priori partitioning of the state space, provides bounds on deciding whether a satisfying trajectory can be found, handles nonconvex \gls*{stl} formulas, and scales well to systems with high-dimensional dynamics. 

We propose a two-layer control architecture with discrete and continuous planning layers based on \gls*{cegis}. These layers work together to iteratively overcome non-convexities in the \gls*{stl} specification. Discrete plans (sequences of convex constraints) are generated on-the-fly from \gls*{stl} predicates, which allows us to avoid expensive conversions between predicates and vertex representations. The discrete layer acts as a \textit{proposer}, using \gls*{smt} to generate a discrete plan which satisfies the \gls*{stl} specification. The continuous layer then acts as a \textit{teacher}, using \gls*{lp} to check if the discrete plan is dynamically feasible. If the discrete plan is not feasible, a counterexample is passed back to the discrete layer, which generates a new plan. If the plan is feasible, the continuous layer returns a dynamically feasible trajectory that satisfies the \gls*{stl} specification.  

\subsection{Related Work}

Existing approaches for symbolic control based on trajectory synthesis from task specifications can be roughly divided into three categories: \gls*{milp} based \cite{wolff2016optimal,raman2014model,sadraddini2015robust,liu2017communication,lindemann2017robust}, sampling based \cite{he2015towards,plaku2016motion},  and \gls*{smt} based \cite{shoukry2016scalable,shoukry2017linear} approaches. 

The basic idea of the \gls*{milp} approach is to re-write statements with logical expressions into mixed-integer constraints. The addition of auxiliary binary variables to facilitate this rewriting, however, may render the problem intractable as for long specifications. Thus, \gls*{stl} approaches such as Blu\gls*{stl} \cite{raman2014model} have been focused on \gls*{mpc} which limits the duration (i.e., number of time indices) of the search. LTL OPT \cite{plaku2016motion} proposed to synthesize controllers from an expressive fragment of \gls*{ltl} specifications for longer time horizons. This approach faces the same limitations, however, and struggles to efficiently handle non-convex logical constraints with long duration (greater than $100$ time indices).          

Sampling-based approaches propose to combine sampling-based motion planning with discrete search algorithms. Sampling-based motion planning methods are relatively easy to implement and can provide fast solutions to some difficult problems. However, such approaches are suboptimal and are not guaranteed to find a solution if one exists, a property referred to as \textit{(in)completeness}. Instead, they ensure weaker notions of \textit{asymptotically optimality} \cite{karaman2011sampling} and \textit{probabilistic completeness} \cite{hsu2006probabilistic}, meaning that an optimal solution will be provided, if one exists, given sufficient runtime of the algorithm. Hence sampling-based algorithms may never terminate and do not have control over the number of hops needed to generate a trajectory. These difficulties are exemplified in poor performance on problems with narrow passages \cite{zhang2008efficient}. 

Our proposed approach is closely related to \gls*{smt} based symbolic control, which has been used to generate dynamically-feasible trajectories for \gls*{ltl} specifications \cite{shoukry2016scalable,shoukry2017linear}. Modern \gls*{smt} solvers can efficiently find satisfying valuations of very large propositional formulas with complex Boolean structure combining various decidable theories, such as lists, arrays, bit vectors, linear integer arithmetic, and linear real arithmetic \cite{barrett2009satisfiability}. \gls*{smt} based symbolic control from \gls*{ltl} specifications showed encouraging performance for motion planning problems. However, it requires partitioning the continuous state space before the execution. This precomputation that can be more time consuming than the trajectory synthesis itself \cite[Table~I]{shoukry2016scalable}. Furthermore, the implementation of real-time specifications is more difficult, and it does not offer an explicit bound to decide that dynamically-feasible trajectory for \gls*{ltl} specifications cannot be found.

\subsection{Contribution}

Our approach explicitly searches for the duration necessary to find a dynamically-feasible trajectory that satisfies the specifications. Moreover, as shown in our experiments, our approach can efficiently handle long duration (greater than $100$ time indices), non-convex task specifications. Unlike sampling-based and other \gls*{smt} approaches, our method guarantees correctness for time-bounded \gls*{stl} specifications and efficiently maximizes the robust satisfaction of these specifications. Furthermore, our method does not require precomputation of an abstract transition system.


\subsection{Organization} 

The rest of the paper is organized as follows. After introducing the necessary preliminaries, Section \ref{sec:preliminaries} presents a formal problem statement. Next, Section \ref{sec:method} outlines our proposed approach to this problem. Sections \ref{sec:dplan}-\ref{sec:robctrler} present the details of our method. Section \ref{sec:results} shows how our method can be used to solve a high-dimensional task and motion planning problem. Section \ref{sec:conclusion} concludes the paper.

\section{Preliminaries}\label{sec:preliminaries}

\subsection{System}\label{sec:system}

Consider the discrete-time linear control system
\begin{equation}\label{eq:system}
\gls*{statec}_{k+1} = A\gls*{statec}_{k} + B\gls*{inputc}_{k}, \quad \gls*{statec}_0 = \gls*{statecini}, 
\end{equation}
where $\gls*{statec} \in \mathbb{R}^{\gls*{statecnb}}$ are the state variables, $\gls*{inputc} \subset \mathbb{R}^{\gls*{inputcnb}}$ are the control inputs, and $A \in \mathbb{R}^{\gls*{statecnb} \times \gls*{statecnb}}$ and $B \in \mathbb{R}^{\gls*{statecnb}\times \gls*{inputcnb}}$ are constant matrices.  
 
Note that System (\ref{eq:system}) can arise from linearization and sampling of a more general continuous system. In this case, we denote the sampling period as $\gls*{ts}$, where $\gls*{ts} = t_{k+1} - t_k $ for all $k \in \mathbb{N}_{\geq 0}$.

A run of System (\ref{eq:system}) is defined as a sequence $\gls*{run}(\gls*{statecini}) = (\gls*{statec}_{0},\gls*{inputc}_{0})(\gls*{statec}_{1},\gls*{inputc}_{1})\dots$ specified by an initial state $\gls*{statecini} \in \mathbb{R}^{\gls*{statecnb}}$ and an input sequence $\gls*{inputc} = \gls*{inputc}_{0}\gls*{inputc}_{1} \dots$. A evaluation of this run at instant $t_k$, $\gls*{run}(t_k) = (\gls*{statec}(t_{k}),\gls*{inputc}(t_{k}))$, is denoted  by $\gls*{run}_k = (\gls*{statec}_{k},\gls*{inputc}_{k})$. 

\subsection{Signal Temporal Logic}\label{sec:stl}

We assume that high-level specifications are given as \gls*{stl} formulas. \gls*{stl} formulas are defined recursively according to the following syntax in conjunctive normal form:
\begin{equation*}
    \gls*{stlformula} := \idstlpred | \gls*{neg} \idstlpred | \gls*{stlformula}_1 \gls*{and} \gls*{stlformula}_2 | \gls*{stlformula}_1 \gls*{or} \gls*{stlformula}_2 | \idstlalways{[a,b]}{\gls*{stlformula}} | \idstlevent{[a,b]}{\gls*{stlformula}} | \idstluntil{\gls*{stlformula}_1}{[a,b]}{\gls*{stlformula}_2},
\end{equation*}
where \gls*{stlformula}, $\gls*{stlformula}_1$, $\gls*{stlformula}_2$  are \gls*{stl} formulas, and $\idstlpred$ is an atomic predicate $\mathbb{R}^{\gls*{statecnb}} \times \mathbb{R}^{\gls*{inputcnb}} \mapsto \{\gls*{true},\gls*{false}\}$. This predicate is determined by the sign of the function $\gls*{stlfunc} : \mathbb{R}^{\gls*{statecnb} + \gls*{inputcnb}} \mapsto \mathbb{R}$. Specifically, $\idstlpred = \gls*{true}$ if and only if $\gls*{stlfunc}(\gls*{run}) > 0$. We assume that $\gls*{stlfunc}$ is linear affine, i.e.,  $\gls*{stlfunc}(\gls*{run}) = \boldsymbol{h}^\intercal \gls*{run} + a$, $\boldsymbol{h} \in \mathbb{R}^{\gls*{statecnb} + \gls*{inputcnb}}$ and $a \in \mathbb{R}$. 

The meaning (semantics) of an \gls*{stl} formula is interpreted over a run of System (\ref{eq:system}).  We denote a run $\gls*{run}(\gls*{statecini})$ satisfying an \gls*{stl} formula \gls*{stlformula} by $\gls*{run} \gls*{sat} \gls*{stlformula}$.  
We write $\gls*{run} \gls*{sat}_k \gls*{stlformula}$ when the run $\gls*{run}_k\gls*{run}_{k+1}\dots$ satisfies the \gls*{stl} formula \gls*{stlformula}. 

The following semantics define the validity of a formula \gls*{stlformula} with respect to the run $\gls*{run}$: 
\begin{itemize}
  \item $\gls*{run} \gls*{sat}_k \idstlpred$ if and only if $\gls*{stlfunc}(\gls*{run}_k) > 0$;
  \item $\gls*{run} \gls*{sat}_k \gls*{neg} \idstlpred$ if and only if $-\gls*{stlfunc}(\gls*{run}_k) > 0$;
  \item $\gls*{run} \gls*{sat}_k \gls*{stlformula}_1 \gls*{and} \gls*{stlformula}_2$ if and only if $\gls*{run} \gls*{sat}_k{\gls*{stlformula}_1}$ and $\gls*{run} \gls*{sat}_k{\gls*{stlformula}_2}$;
  \item $\gls*{run} \gls*{sat}_k \gls*{stlformula}_1 \gls*{or} \gls*{stlformula}_2$ if and only if $\gls*{run} \gls*{sat}_k{\gls*{stlformula}_1}$ or $\gls*{run} \gls*{sat}_k{\gls*{stlformula}_2}$;
  \item $\gls*{run} \gls*{sat}_k \idstlalways{[a,b]}{\gls*{stlformula}}$ if and only if $\forall t_{k^\prime} \in [t_k+a,t_k+b] \, \text{ s.t. } \gls*{run} \gls*{sat}_{k^\prime}{\gls*{stlformula}}$;
  \item $\gls*{run} \gls*{sat}_k \idstlevent{[a,b]}{\gls*{stlformula}}$ if and only if $\exists t_{k^\prime} \in [t_k+a,t_k+b] \,  \text{ s.t. } \gls*{run} \gls*{sat}_{k^\prime}{\gls*{stlformula}}$;
  \item $\gls*{run} \gls*{sat}_k \idstluntil{\gls*{stlformula}_1}{[a,b]}{\gls*{stlformula}_2}$ if and only if $\exists t_{k^\prime} \in [t_k+a,t_k+b]$ s.t. $\gls*{run} \gls*{sat}_{k^\prime}{\gls*{stlformula}_2}$, and $\gls*{run} \gls*{sat}_{k^{\prime\prime}}{\gls*{stlformula}_1}$ for any $t_{k^{\prime\prime}} \in [t_k..t_{k^\prime}]$;
  \item $\gls*{run} \gls*{sat} \gls*{stlformula}$ if and only if $\gls*{run} \gls*{sat}_0 \gls*{stlformula}$. 
\end{itemize} 
If an \gls*{stl} formula \gls*{stlformula} contains no unbounded operators, it is called \textit{bounded-time}. The \textit{bound} of a bounded-time \gls*{stl} formula \gls*{stlformula} is the maximum over the sum of all nested upper bounds on the temporal operators, which is a conservative maximum time length required to decide its satisfiability. For instance, the bounded-time formula $\gls*{stlformula} = \idstlalways{[0,180]}{\idstlevent{[0,90]}{x>0}}$ has a bound $\gls*{kmax} \leq 180 + 90 = 270$, where the bound \gls*{kmax} is sufficient to determine whether the formula is satisfiable. 

\subsection{Quantitative Semantics for \glsfmtshort{stl}}\label{sec:robstl}

Quantitative (or robust) semantics for \gls*{stl} defines a real-valued function $\idstlrobfunc{\gls*{stlformula}}$ called \textit{robustness}, which associates a scalar value with the quality of satisfaction. This robustness function is recurslively defined such that $\idstlrobfunc[0]{\gls*{stlformula}} > 0$ only if $\gls*{run} \gls*{sat} \gls*{stlformula}$:

\begin{align*}
\idstlrobfunc{\idstlpred}(\gls*{run},t_k) := & \gls*{stlfunc}(\gls*{run}_k), \\
\idstlrobfunc{\gls*{neg} \idstlpred}(\gls*{run},t_k) := & -\idstlrobfunc{\idstlpred}(\gls*{run},t_k), \\
\idstlrobfunc{\gls*{stlformula}_1 \gls*{and} \gls*{stlformula}_2}(\gls*{run},t_k) := & \min(\idstlrobfunc{\gls*{stlformula}_1}(\gls*{run},t_k),\idstlrobfunc{\gls*{stlformula}_2}(\gls*{run},t_k)), \\
\idstlrobfunc{\gls*{stlformula}_1 \gls*{or} \gls*{stlformula}_2}(\gls*{run},t_k) := & \max(\idstlrobfunc{\gls*{stlformula}_1}(\gls*{run},t_k),\idstlrobfunc{\gls*{stlformula}_2}(\gls*{run},t_k)) \\
\idstlrobfunc{\idstlalways{[a,b]}{\gls*{stlformula}}}(\gls*{run},t_k) := & \min_{t_{k^\prime} \in [t_k+a,t_k+b]} \idstlrobfunc{\gls*{stlformula}}(\gls*{run},t_{k^\prime}), \\
\idstlrobfunc{\idstlevent{[a,b]}{\gls*{stlformula}}}(\gls*{run},t_k) := & \max_{t_{k^\prime} \in [t_k+a,t_k+b]} \idstlrobfunc{\gls*{stlformula}}(\gls*{run},t_{k^\prime}), \\
\idstlrobfunc{\idstluntil{\gls*{stlformula}_1}{[a,b]}{\gls*{stlformula}_2}}(\gls*{run},t_k) := & \max_{t_{k^\prime} \in [t_k+a,t_k+b]}\Big(\min\big( \idstlrobfunc{\gls*{stlformula}_2}(\gls*{run},t_{k^\prime}), \hspace{1cm} \\
& \pushright{\min_{t_{k^{\prime\prime}} \in [t_k,t_{k^\prime}]} \idstlrobfunc{\gls*{stlformula}_1}(\gls*{run},t_{k^{\prime\prime}})\big)\Big),}
\end{align*}
where we denote $\idstlrobfunc{\gls*{stlformula}}(\gls*{run},t_k)$ by $\idstlrobfunc[k]{\gls*{stlformula}}$ to simplify notation. The robust satisfaction of \gls*{stl} formula \gls*{stlformula} is computed by propagating the values of the functions associated with each operand using $\min$ and $\max$ operators. 

\subsection{Problem formulation}\label{sec:problem}

The \gls*{stl} symbolic control problem is formally defined as follows:

\begin{problem}\label{prob:1}
Given an \gls*{stl} formula \gls*{stlformula} and a dynamical system (\ref{eq:system}) with an initial state $\gls*{statecini} \in \mathbb{R}^{\gls*{statecnb}}$, design control signals $\gls*{inputc} = \gls*{inputc}_{0}\gls*{inputc}_{1}\dots$ such that the corresponding run 
$\gls*{run}(\gls*{statecini}) := (\gls*{inputc}_{0}\gls*{statec}_{0})(\gls*{inputc}_{1}\gls*{statec}_{1}), \dots$  satisfies \gls*{stlformula} and the dynamical constraints of System (\ref{eq:system}). 
\end{problem}

\subsection{Overview}\label{sec:method}

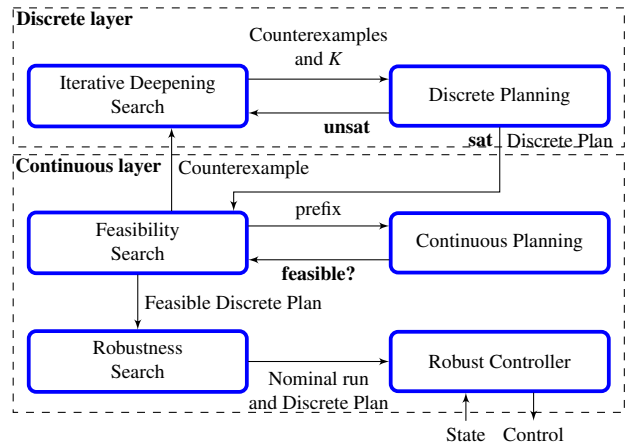
\begin{figure}
\tikzstyle{block} = [draw, rounded corners=1mm, color=blue, text=black, line width=0.5mm, rectangle, minimum height=3em, minimum width=11em]
\tikzstyle{block2} = [draw, color=black, text=black, line width=0.5mm, rectangle, minimum height=3em, minimum width=11em]
\centering
\begin{tikzpicture}[auto, >=latex', scale=0.75, transform shape]	
\node[block] (dplan) {Discrete Planning};
\node[block,below=1.5cm of dplan] (cplan) {Continuous Planning};
\node[block,below=1cm of cplan] (ctrler) {Robust Controller};
\node[block,left=2.5cm of dplan] (idsearch) {\begin{tabular}{c}Iterative Deepening \\ Search\end{tabular} };
\node[block,left=2.5cm of cplan] (fsearch) {\begin{tabular}{c}Feasibility \\ Search\end{tabular} };
\node[block,left=2.5cm of ctrler] (rsearch) {\begin{tabular}{c}Robustness \\ Search\end{tabular} };
\node[above left = 0.5cm and -2cm of idsearch] {\textbf{Discrete layer}};
\node[above left = 0.5cm and -2.5cm of fsearch] {\textbf{Continuous layer}};
\draw[->] ([yshift=3mm] idsearch.east) -- node[midway,above] {\begin{tabular}{c} Counterexamples \\ and $K$ \end{tabular}} ([yshift=3mm] dplan.west);
\draw[<-] ([yshift=-3mm] idsearch.east) -- node[pos=0.7,below] {\textbf{unsat}} ([yshift=-3mm] dplan.west);
\draw[->] (dplan.south) -- node[pos=0.2,right] {Discrete Plan} ++(0cm,-1.15cm) node[pos=0.2,left] {\textbf{sat}} -| ([xshift=17mm] fsearch.north);
\draw[->] ([xshift=6mm] fsearch.north) -- node[midway,right] {Counterexample} ([xshift=6mm] idsearch.south);
\draw[->] ([yshift=3mm] fsearch.east) -- node[midway,above] {prefix} ([yshift=3mm] cplan.west);
\draw[<-] ([yshift=-3mm] fsearch.east) -- node[midway,below] {\textbf{feasible?}} ([yshift=-3mm] cplan.west);
\draw[->] (fsearch.south) -- node[midway,right] {Feasible Discrete Plan} (rsearch.north);
\draw[->] (rsearch.east) -- node[midway,below] {\begin{tabular}{c}Nominal run \\ and Discrete Plan \end{tabular}} (ctrler.west);
\draw[->] ([xshift=6mm] ctrler.south) -- node[pos=1,below] {Control} ++(0,-0.5cm);
\draw[<-] ([xshift=-6mm] ctrler.south) -- node[pos=1,below] {State} ++(0,-0.5cm);
\draw[-,dashed] ([xshift=-0.25cm,yshift=1cm] idsearch.north -| idsearch.west) -| ([xshift=0.25cm,yshift=-0.35cm] dplan.south -| dplan.east) -| cycle;
\draw[-,dashed] ([xshift=-0.25cm,yshift=1cm] fsearch.north -| fsearch.west) -| ([xshift=0.25cm,yshift=-0.35cm] ctrler.south -| ctrler.east) -| cycle;
\end{tikzpicture}
\caption{Pictorial representation of proposed approach. }
\label{fig:diag1}
\end{figure}

We achieve scalable symbolic control by applying search techniques to reduce the number of discrete and continuous planning computations. Our approach is illustrated in Fig. \ref{fig:diag1}. At discrete planning layer, we combine iterative deepening search \cite{kira:Russell:2009} with a \gls*{csp} \cite{de2011satisfiability}. Since \gls*{csp} generates finite length solutions, we search for minimum length solutions for the discrete planning problem by iteractively incrementing constraints. At the continuous planning layer, we use the fact that if there exists a prefix of the discrete plan that is not dynamically feasible, finding a dynamically feasible suffix will not change the feasibility of the prefix. This combination of best-first and binary search with optimization to generate discrete plans ensures that computation will focus on plans that are likely to be dynamically feasible. Similarly, we avoid computing new discrete plans until we check the dynamic feasibility of all related plans. Leveraging these aggressive search techniques with the efficiency of \gls*{smt} solvers, our approach extends symbolic control scalability to more complex task specifications and higher dimensional systems. 

\begin{figure}
\tikzstyle{block} = [draw, fill=blue!20, rectangle, minimum height=3em, minimum width=6em]
\tikzstyle{pinstyle} = [pin edge={to-,thin,black}]
\tikzstyle{sum} = [draw, fill=blue!20, circle, node distance=1cm]
\centering
\begin{tikzpicture}[auto, >=latex', scale=0.75, transform shape]
    \node [block] (dplan1) {\begin{tabular}{c} Discrete \\ Planning \end{tabular}};    	    
    \node [block, right =2cm of dplan1] (cplan1) {\begin{tabular}{c} Continuous \\ Planning \end{tabular}};       
    \node [above left=0.25cm and 0.25cm of dplan1] (K1) {K=0};
    \node [below right=0.15cm and 0.25cm of cplan1] (out1) {$\gls*{polyseq}, L$};
    \node [below left=0cm and -1.25cm of cplan1] {\textbf{feas?}};

    \node [block, below=1.25cm of dplan1] (dplan2) {\begin{tabular}{c} Discrete \\ Planning \end{tabular}};    	    
    \node [block, right =2cm of dplan2] (cplan2) {\begin{tabular}{c} Continuous \\ Planning \end{tabular}};       
    \node [below right=0.15cm and 0.25cm of cplan2] (out2) {$\gls*{polyseq}, L$};
    \node [below left=-0.35cm and 0.25cm of dplan1] {\textbf{unsat?}};
    \node [above left=-0.15cm and 0.3cm of dplan2] {$K=K+1$};
    \node [above left=0.4cm and -2.5cm of dplan2] {\color{blue} deepening iteration};
    \node [below left=0cm and -1.25cm of cplan2] {\textbf{feas?}};

    \node [block, below=1.25cm of dplan2] (dplanmax) {\begin{tabular}{c} Discrete \\ Planning \end{tabular}};    	    
    \node [block, right =2cm of dplanmax] (cplanmax) {\begin{tabular}{c} Continuous \\ Planning \end{tabular}};       
    \node [below right=0.15cm and 0.25cm of cplanmax] (outmax) {$\gls*{polyseq}, L$};
    \node [below left=-0.35cm and 0.25cm of dplan2] {\textbf{unsat?}};
    \node [above left=-0.15cm and 0.3cm of dplanmax] {$K=\gls*{kmax}$};
    \node [below left=0cm and -1.25cm of cplanmax] {\textbf{feas?}};
    \node [below=0.2cm of dplan2] {$\textbf{\vdots}$};
    \coordinate [below left=0.2cm and 0.35cm of dplan2] (unsat2);
    \coordinate [above left=0.2cm and 0.35cm of dplanmax] (kmax);
    \node [below left=-0.35cm and 0.25cm of dplanmax] {\textbf{unsat?}};
    \node [below left=0.6cm and -5cm of dplanmax] {No dynamically feasible and satisfying run exists.};
    \coordinate [below left=0.5cm and 0.35cm of dplanmax] (unsatmax);

    \draw[->] (K1.south) to [out=-90,in=180] ([yshift=0.35cm] dplan1.west);
    \draw[->] (cplan1.south) to [out=-90,in=180] (out1.west);
     \path[pil]
 	([yshift=-5mm] cplan1.west) edge node[auto] {$\gls*{polyseq}_{cex}$} node[near start,above] {\textbf{infeas?}} ([yshift=-5mm] dplan1.east)
    ([yshift=5mm] dplan1.east) edge node[auto] {$\gls*{poly}_0$}  node[near start,below] {\textbf{sat?}} ([yshift=5mm] cplan1.west);
    \draw[->] ([yshift=-0.35cm] dplan1.west) to [out=-180,in=180] ([yshift=0.35cm] dplan2.west);
    \draw[->] (cplan2.south) to [out=-90,in=180] (out2.west);
     \path[pil]
 	([yshift=-5mm] cplan2.west) edge node[auto] {$\gls*{polyseq}_{cex}$} node[near start,above] {\textbf{infeas?}} ([yshift=-5mm] dplan2.east)
    ([yshift=5mm] dplan2.east) edge node[auto] {$\gls*{poly}_0\gls*{poly}_1$}  node[near start,below] {\textbf{sat?}} ([yshift=5mm] cplan2.west);
    \draw[->] ([yshift=-0.35cm] dplan2.west) to [out=-180,in=90] (unsat2);
    \draw[->] (kmax) to [out=-90,in=150] ([yshift=0.35cm] dplanmax.west);
    \draw[->] (cplanmax.south) to [out=-90,in=180] (outmax.west);
     \path[pil]
 	([yshift=-5mm] cplanmax.west) edge node[auto] {$\gls*{polyseq}_{cex}$} node[near start,above] {\textbf{infeas?}} ([yshift=-5mm] dplanmax.east)
    ([yshift=5mm] dplanmax.east) edge node[auto] {$\gls*{poly}_0\gls*{poly}_1\dots\gls*{poly}_{\gls*{kmax}}$}  node[near start,below] {\textbf{sat?}} ([yshift=5mm] cplanmax.west);
    \draw[->] ([yshift=-0.35cm] dplanmax.west) to [out=-180,in=90] (unsatmax);
\end{tikzpicture}
\caption{Pictorial representation of idSTL Algorithm.}
\label{fig:idsearch}
\end{figure}
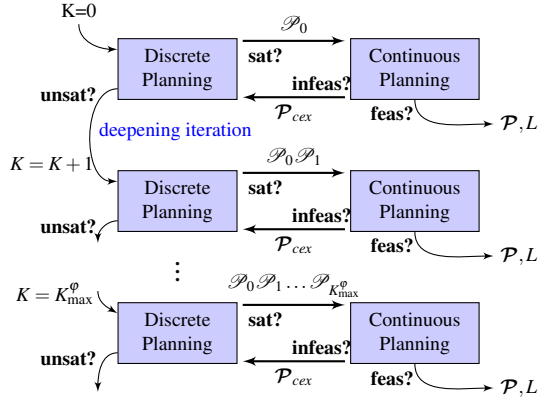

Algorithm \ref{algo:idstl} (illustrated in Fig. \ref{fig:idsearch}) outlines our approach. The main loop implements the iterative deepening search. We start with discrete plans of length $K=0$ (line \ref{alg:idstl_init}). If the discrete planning is not satisfiable with this length, then we deepen the discrete planning problem by iterating $K$ (line \ref{alg:idstl_deep}). If the discrete planning problem is satisfiable for a given length $K$, we start a search for a dynamically feasible run with minimum length ($feas$ algorithm). Instead of asking for new discrete plans, this algorithm modifies discrete plan prefixes to propose new candidate plans. If we do not find a dynamically feasible plan, this algorithm identifies prefixes that are infeasible and must be discarded in next searches. If we find a feasible discrete plan, we increase its \gls*{stl} robustness measure $\idstlrobfunc{\gls*{stlformula}}$ with the robustness search algorithm ($rob$). This last search will generate a robust nominal run and discrete plan, which is used with a feedback regulator synthesized in  algorithm $ctrler$. In the following sections, we will present each one of these algorithms in detail.    
 
\begin{algorithm}
	\caption{idSTL}\label{algo:idstl}
    \begin{algorithmic}[1]
		\REQUIRE \idstlinput
		\ENSURE \idstloutput
			\STATE $K \gls*{assert} 0$, $\quad \text{Status} \gls*{assert} \text{unsat}$, $\quad \text{isFeasible} \gls*{assert} \gls*{false}$, $\quad \gls*{polyseq}_{cex} \gls*{assert} \emptyset$ \label{alg:idstl_init}
            \WHILE{$K \leq \gls*{kmax} \gls*{and} \big( \text{Status} = \text{unsat} \gls*{or} \gls*{neg} \text{isFeasible} \big)$}
                \STATE $\idstldplanoutput \gls*{assert} dplan\idstldplaninput$
				\STATE $\gls*{polyseq}_{cex} \gls*{assert} \emptyset$
                \IF{$\text{Status} = \text{sat}$}                 
					\STATE $\idstlfeasoutput \gls*{assert} feas\idstlfeasinput$
                \ELSE 
    	            \STATE $K \gls*{assert} K + 1$ \label{alg:idstl_deep}
                \ENDIF
			\ENDWHILE			
			\STATE $\idstlroboutput \gls*{assert} rob\idstlrobinput$
  \end{algorithmic}
\end{algorithm}

\section{Discrete Planning}\label{sec:dplan}

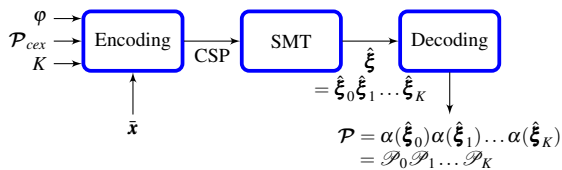
\begin{figure}
\tikzstyle{block} = [draw, rounded corners=1mm, color=blue, text=black, line width=0.5mm, rectangle, minimum height=3em, minimum width=5em]
\tikzstyle{block2} = [draw, color=black, text=black, line width=0.5mm, rectangle, minimum height=3em, minimum width=9em]
\centering
\begin{tikzpicture}[auto, >=latex', scale=0.75, transform shape]	
\node[block] (enc) {Encoding};
\node[block,right=1cm of enc] (smt) {SMT};
\node[block,right=1cm of smt] (dec) {Decoding};
\draw[->] (enc.east) -- node[below,midway] {CSP} (smt.west); 
\draw[->] (smt.east) -- node[below,midway] {\begin{tabular}{c} \gls*{run_coarse} \\ $ = \gls*{run_coarse}_0\gls*{run_coarse}_1\dots\gls*{run_coarse}_K$ \end{tabular}} (dec.west); 
\draw[->] (dec.south) -- node[below,pos=1] {\begin{tabular}{l} $\gls*{polyseq} = \gls*{abs}(\gls*{run_coarse}_0)\gls*{abs}(\gls*{run_coarse}_1)\dots\gls*{abs}(\gls*{run_coarse}_K)$ \\ \hspace{0.3cm} $= \gls*{poly}_0\gls*{poly}_1\dots\gls*{poly}_K$ \end{tabular}} ++(0,-0.75cm); 
\draw[<-] ([yshift= 4mm] enc.west) -- node[pos=1,left] {\gls*{stlformula}} ++(-0.5cm,0);
\draw[<-] ([yshift= 0mm] enc.west) -- node[pos=1,left] {$\gls*{polyseq}_{cex}$} ++(-0.5cm,0);
\draw[<-] ([yshift=-4mm] enc.west) -- node[pos=1,left] {$K$} ++(-0.5cm,0);
\draw[<-] ([yshift= 0mm] enc.south) -- node[pos=1,below] {$\gls*{statecini}$} ++(0cm,-0.75cm);
\end{tikzpicture}
\caption{Graphical description of proposed discrete planning. }
\label{fig:diag2}
\end{figure}

We encode the discrete planning problem into a \gls*{csp} \cite{de2011satisfiability}, allowing us to leverage \textit{off-the-shelf} incremental \gls*{smt} solvers such as Z3 \cite{de2008z3}. The \gls*{csp} is mathematical problem, essentially specifying that a set of states must satisfy a set of constraints. \glspl*{csp} with constraints defined in quantifier-free linear first-order logic can be efficiently (polynomial time) computed using \gls*{smt} solvers \cite{barrett2009satisfiability}. Since \glspl*{csp} are defined over finite sets of variables, however, we can only check satisfiability of runs with finite length. As shown in Fig. \ref{fig:diag2}, the discrete planner starts by encoding an \gls*{stl} formula \gls*{stlformula} into a \gls*{csp} to generate a candidate run \gls*{run_coarse} with length $K$, discarding any solution with a prefix in the counterexamples $\gls*{polyseq}_{cex}$. Given such a candidate run \gls*{run_coarse}, we apply an abstraction operator \gls*{abs} to generate a discrete plan \gls*{polyseq}. 

\subsection{Problem Formulation} 

To scaleably obtain a discrete plan without a pre-partitioned workspace, we generate a run as sequence of closed convex polyhedra \gls*{polyseq} from an automatically synthesized coarse run \gls*{run_coarse}. This allows us to avoid computing partitions from a set of vertices, which requires prohibitively expensive conversions between halfspace resulting from \gls*{stl} predicates \idstlpred and vertex representations.

To show the relationship between a coarse run \gls*{run_coarse} and a run of system (\ref{eq:system}), we model the system as a transition system \gls*{tsys} and propose an existential abstraction \gls*{tsys}$^\prime$ that explains why our discrete plan over-approximates the continuous plan. Note that we never explicitly use the existential abstraction transition system to generate discrete plans. 

\begin{definition}\label{def:model}
Given an initial state \gls*{statecini} and an \gls*{stl} formula \gls*{stlformula}, $\gls*{tsys} = \langle \mathcal{S}, \mathcal{S}_0, \mathcal{L}, \delta \rangle$ is a labeled transition system where,
\begin{itemize}
  \item $\mathcal{S} := \mathbb{R}^{\gls*{statecnb}+\gls*{inputcnb}}$ is the transition system state continuous domain, where $\gls*{run}_k = (\gls*{statec}_k,\gls*{inputc}_k)$ is a state $\gls*{run}_k \in \mathcal{S}$ at $t_k$,
  \item $\mathcal{S}_0 \subseteq \mathcal{S}$ is the initial conditions such that for any $\gls*{run}_0 \in \mathcal{S}_0$, $\gls*{run}_0 = (\gls*{statecini},\gls*{inputc})$ for any $\gls*{inputc} \in \mathbb{R}^{\gls*{inputcnb}}$,
  \item $\mathcal{L} : \mathcal{S} \mapsto 2^{\idstlpred}$\footnote{$2^{\mathcal{A}}$ is the power set of the domain $\mathcal{A}$.} is label function mapping a state $\gls*{run}_k$ to all predicates defined in the \gls*{stl} formula \gls*{stlformula} such that $\idstlpred \in \mathcal{L}(\gls*{run}_k)$ if and only if $\gls*{run} \gls*{sat}_k \idstlpred$,     
  \item $\delta$ is a transition relation where $(\gls*{run}_k,\gls*{run}_{k+1}) \in \delta$ if and only if $\gls*{statec}_{k+1} = A\gls*{statec}_k + B\gls*{inputc}_k$, and $\gls*{run}_k$ and $ \gls*{run}_{k+1}$ are adjacent. It means that there exists a $\idstlpred \in \mathcal{L}(\gls*{run}_k) \cap \mathcal{L}(\gls*{run}_{k+1})$.
\end{itemize}
\end{definition}
It easy to see that a run \gls*{run} is a run of \gls*{tsys} only if it is a run of a system (\ref{eq:system}). Note that we include the adjacency constraint used in coarse abstraction approaches like \cite{he2015towards,plaku2016motion,shoukry2016scalable,shoukry2017linear}. We do so because though we are using a difference equation to model our system, the underlying system is continuous. Thus, depending on the sampling time $\gls*{ts}$, the difference equation may allow transitions which may exist an instant $t_{k^\prime} \in [t_{k},t_{k+1}]$ such that $\mathcal{L}(\gls*{run}_{k^\prime}) \cap \mathcal{L}(\gls*{run}_k) = \gls*{emptyset}$ and $\mathcal{L}(\gls*{run}_{k^\prime}) \cap \mathcal{L}(\gls*{run}_{k+1}) = \gls*{emptyset}$, violating the \gls*{stl} formula \gls*{stlformula}.  

The existential abstraction \gls*{tsys}$^\prime$ is formally defined as follows:
\begin{definition}\label{def:eabsmodel}
Given an initial state \gls*{statecini} and an \gls*{stl} formula \gls*{stlformula}, $\gls*{tsys}^\prime = \langle \mathcal{S}^\prime, \mathcal{S}_0^\prime, \mathcal{L}^\prime, \delta^\prime \rangle$ is a labeled transition system where,
\begin{itemize}
  \item $\mathcal{S}^\prime := \{ \gls*{poly}^{[1]},\dots,\gls*{poly}^{[N]} \}$ is a finite set of discrete states representing all possible convex polyhedra generated by the predicates of the \gls*{stl} formula \gls*{stlformula},
  \item $\mathcal{S}_0^\prime \subseteq \mathcal{S}$ is the set of initial conditions such that $\gls*{poly}_0 \in \mathcal{S}_0^\prime$ only if $(\gls*{statecini},\gls*{inputc}) \in \gls*{poly}_0$ for some $\gls*{inputc} \in \mathbb{R}^{\gls*{inputcnb}}$,
  \item $\mathcal{L}^\prime : \mathcal{S}^\prime \mapsto 2^{\idstlpred}$ is label function mapping a state $\gls*{poly}_k$ to all predicates defined in the \gls*{stl} formula \gls*{stlformula}, where $\idstlpred \in \mathcal{L}(\gls*{poly}_k)$ if and only if $\gls*{run} \gls*{sat}_k \idstlpred$ for any run $\gls*{run} \in (\mathbb{R}^{\gls*{statecnb}+\gls*{inputcnb}})^\omega$ such that $\gls*{run}_k \in \gls*{poly}_k$,     
  \item $\delta^\prime$ is a transition relation where $(\gls*{poly}_k,\gls*{poly}_{k+1}) \in \delta^\prime$ if and only if there exist $\gls*{poly} \in \mathcal{S}^\prime$ such that $\gls*{poly}_k \cap \gls*{poly} \neq \gls*{emptyset}$ and $\gls*{poly}_{k+1} \cap \gls*{poly} \neq \gls*{emptyset}$.
\end{itemize}
\end{definition}
A run of model \gls*{tsys}$^\prime$ is a sequence $\gls*{polyseq} = \gls*{poly}_0\gls*{poly}_1\gls*{poly}_2\dots$. Since an \gls*{stl} formula \gls*{stlformula} has a finite set of predicates by definition, the number of all possible convex polyhedra is finite; thus, $\mathcal{S}^\prime$ is finite. 

\begin{example}
Consider transitions illustrated in Fig.~\ref{fig:ex1_transition}. All convex polygons $\gls*{poly}^{[i]}$ are states of \gls*{tsys}$^\prime$, as are all combinations of adjacent polygons. The transition $(\gls*{poly}^{[1]},\gls*{poly}^{[5]})$ is valid (i.e., $(\gls*{poly}^{[1]},\gls*{poly}^{[5]}) \in \delta^\prime$) as shown in Fig.~\ref{fig:ex1_transition01}, because the polygon $\gls*{poly}^{[1,3,5]} = \gls*{poly}^{[1]} \cup \gls*{poly}^{[3]} \cup \gls*{poly}^{[5]}$, i.e., the red rectangle in Fig.~\ref{fig:ex1_transition01}, is also a state of \gls*{tsys}$^\prime$. However, $(\gls*{poly}^{[1]},\gls*{poly}^{[8]})$, shown in Fig.~\ref{fig:ex1_transition02} is not valid. Also, we can see that $\mathcal{L}^\prime(\gls*{poly}^{[1]}) = \{ \idstlpred[x > 0], \idstlpred[10-x > 0], \idstlpred[y > 0], \idstlpred[10-y > 0]\}$.
 
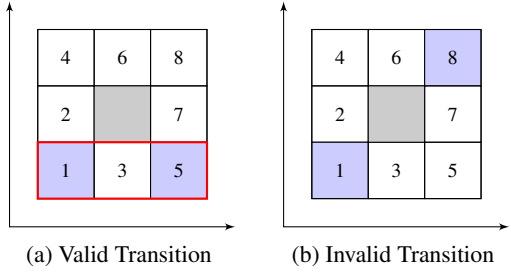
\begin{figure}
\centering
\begin{subfigure}[b]{0.2\textwidth}
\centering
\raisebox{-0.5\height}{
\begin{tikzpicture}[auto, >=latex', scale=0.75, transform shape,level1/.style={sibling distance=50mm,level distance=1cm},level2/.style={sibling distance=10mm,level distance=1cm},level3/.style={sibling distance=30mm,level distance=1cm},level4/.style={sibling distance=15mm,level distance=1cm}]
   \draw[draw=black ] (0cm,0cm) -| ++(1cm,-1cm) -| cycle;
   \draw[draw=black ] (1cm,0cm) -| ++(1cm,-1cm) -| cycle;
   \draw[draw=black ] (2cm,0cm) -| ++(1cm,-1cm) -| cycle;
   \draw[draw=black ] (0cm,-1cm) -| ++(1cm,-1cm) -| cycle;
   \draw[draw=black ] (2cm,-1cm) -| ++(1cm,-1cm) -| cycle;
   \draw[draw=black,fill=blue!20 ] (0cm,-2cm) -| ++(1cm,-1cm) -| cycle;
   \draw[draw=black ] (1cm,-2cm) -| ++(1cm,-1cm) -| cycle;
   \draw[draw=black,fill=blue!20 ] (2cm,-2cm) -| ++(1cm,-1cm) -| cycle;
   \draw[draw=black,fill=black!20 ] (1cm,-1cm) -| ++(1cm,-1cm) -| cycle;
   \draw[draw=black] (0cm,0cm) -| ++(3cm,-3cm) -| cycle;
   \draw[draw=red,thick] (0cm,-2cm) -| ++(3cm,-1cm) -| cycle;
   \draw[<->] (-0.5,0.5) |- (3.5cm,-3.5cm);
   \node at (0.5cm,-0.5cm) {$4$};
   \node at (1.5cm,-0.5cm) {$6$};
   \node at (2.5cm,-0.5cm) {$8$};
   \node at (0.5cm,-1.5cm) {$2$};
   \node at (2.5cm,-1.5cm) {$7$};
   \node at (0.5cm,-2.5cm) {$1$};
   \node at (1.5cm,-2.5cm) {$3$};
   \node at (2.5cm,-2.5cm) {$5$};
\end{tikzpicture}}
\caption{Valid Transition}
\label{fig:ex1_transition01}
\end{subfigure}
\begin{subfigure}[b]{0.20\textwidth}
\centering
\raisebox{-0.5\height}{
\begin{tikzpicture}[auto, >=latex', scale=0.75, transform shape,level1/.style={sibling distance=50mm,level distance=1cm},level2/.style={sibling distance=10mm,level distance=1cm},level3/.style={sibling distance=30mm,level distance=1cm},level4/.style={sibling distance=15mm,level distance=1cm}]
   \draw[draw=black ] (0cm,0cm) -| ++(1cm,-1cm) -| cycle;
   \draw[draw=black ] (1cm,0cm) -| ++(1cm,-1cm) -| cycle;
   \draw[draw=black,fill=blue!20 ] (2cm,0cm) -| ++(1cm,-1cm) -| cycle;
   \draw[draw=black ] (0cm,-1cm) -| ++(1cm,-1cm) -| cycle;
   \draw[draw=black ] (2cm,-1cm) -| ++(1cm,-1cm) -| cycle;
   \draw[draw=black,fill=blue!20 ] (0cm,-2cm) -| ++(1cm,-1cm) -| cycle;
   \draw[draw=black ] (1cm,-2cm) -| ++(1cm,-1cm) -| cycle;
   \draw[draw=black ] (2cm,-2cm) -| ++(1cm,-1cm) -| cycle;
   \draw[draw=black,fill=black!20 ] (1cm,-1cm) -| ++(1cm,-1cm) -| cycle;
   \draw[draw=black] (0cm,0cm) -| ++(3cm,-3cm) -| cycle;
   \draw[<->] (-0.5,0.5) |- (3.5cm,-3.5cm);
   \node at (0.5cm,-0.5cm) {$4$};
   \node at (1.5cm,-0.5cm) {$6$};
   \node at (2.5cm,-0.5cm) {$8$};
   \node at (0.5cm,-1.5cm) {$2$};
   \node at (2.5cm,-1.5cm) {$7$};
   \node at (0.5cm,-2.5cm) {$1$};
   \node at (1.5cm,-2.5cm) {$3$};
   \node at (2.5cm,-2.5cm) {$5$};
\end{tikzpicture}}
\caption{Invalid Transition}
\label{fig:ex1_transition02}
\end{subfigure}
\caption{Illustrative example of transitions in the abstracted system.}
\label{fig:ex1_transition}
\end{figure}

\end{example}

To verify that the abstraction of every run in \gls*{tsys} corresponds to a run in \gls*{tsys}$^\prime$, we propose the following abstraction operator: 
\begin{definition}\label{def:eabsmodel}
$\gls*{abs} : \mathcal{S} \mapsto \mathcal{S}^\prime$ is a function that maps a state $\gls*{run}_k \in \mathcal{S}$ of model \gls*{tsys} to a state $\gls*{poly}_k \in \mathcal{S}^\prime$ of model \gls*{tsys}$^\prime$ as follows,
\begin{equation}
\gls*{poly}_k = \gls*{abs}(\gls*{run}_k) := \{ \gls*{run}_k^\prime \in \mathbb{R}^{\gls*{statecnb}+\gls*{inputcnb}}: \bigwedge_{\forall \idstlpred \in \mathcal{L}(\gls*{run}_k)} \gls*{stlfunc}(\gls*{run}_k^\prime) > 0 \}. 
\end{equation}
\end{definition}

Intuitively, \gls*{tsys}$^\prime$ over-approximates \gls*{tsys}. 
\begin{theorem}\label{teo:eabs}
The abstraction of every run \gls*{run} in \gls*{tsys} is a run in \gls*{tsys}$^\prime$, i.e. $\gls*{run} = \gls*{run}_0\gls*{run}_1\dots \in \gls*{tsys}$ only if $\gls*{polyseq} = \gls*{abs}(\gls*{run}_0)\gls*{abs}(\gls*{run}_1)\dots \in \gls*{tsys}^\prime$.
\end{theorem}
\begin{proof}
By definition \cite{baier2008principles}, $\Sigma^\prime$ simulates $\Sigma$ if and only if: (1) $\mathcal{L}(\gls*{run}_k) = \mathcal{L}\big(\gls*{abs}(\gls*{run}_k)\big)$, (2) $\gls*{run}_0 \in \mathcal{S}_0$ only if $\gls*{abs}(\gls*{run}_0) \in \mathcal{S}_0^\prime$, and (3) $(\gls*{run}_k,\gls*{run}_{k+1}) \in \delta$ only if $(\gls*{abs}(\gls*{run}_k),\gls*{abs}(\gls*{run}_{k+1})) \in \delta^\prime$. 

(1) By definition, a state $\gls*{poly}_k$ is formed by conjunction of inequalities $\gls*{stlfunc}(\gls*{run}_k^\prime) > 0$ such that $\idstlpred \in \mathcal{L}(\gls*{run}_k)$. Therefore, $\gls*{run} \gls*{sat}_k^{(K,L)} \idstlpred$ for any run $\gls*{run} \in (\mathbb{R}^{\gls*{statecnb}+\gls*{inputcnb}})^\omega$ such that $\gls*{run}_k \in \gls*{poly}_k$.     

(2) Since $\gls*{run}_0 = (\gls*{statecini},\gls*{inputc})$ and $\idstlpred \in \mathcal{L}(\gls*{run}_0)$ means that $\gls*{run} \gls*{sat}_0^{(K,L)} \idstlpred$, thus, $\gls*{poly}_0 = \gls*{abs}(\gls*{run}_0)$ is a convex polyhedron which contains $\gls*{statecini}$. Hence, $\gls*{poly}_0 \in \mathcal{S}_0^\prime$. 

(3) Note that the transition relation $(\gls*{run}_k,\gls*{run}_{k+1}) \in \delta$ requires the adjacency constraint, i.e., there exists a $\idstlpred[\gls*{stlfunc}^\prime] \in \mathcal{L}(\gls*{run}_k)$ such that $\mathcal{L}(\gls*{run}_k) / \idstlpred[\gls*{stlfunc}^\prime] = \mathcal{L}(\gls*{run}_{k+1}) / \idstlpred[\gls*{stlfunc}^\prime]$. Hence, there exist a polyhedron $\gls*{poly} = \{ \gls*{run} \in \mathbb{R}^{\gls*{statecnb} + \gls*{inputcnb}} : \bigwedge_{\forall \idstlpred \in \mathcal{L}(\gls*{run}_k) / \idstlpred} \gls*{stlfunc}(\gls*{run}) > 0 \}$ such that $\gls*{poly}_k \cap \gls*{poly} \neq \gls*{emptyset}$ and $\gls*{poly}_{k+1} \cap \gls*{poly} \neq \gls*{emptyset}$. Therefore, $\big(\gls*{abs}(\gls*{run}_k),\gls*{abs}(\gls*{run}_{k+1})\big) \in \delta^\prime$.

Finally, from \cite[Lemma 7.55]{baier2008principles}, it follows that every run $\gls*{run}$ in $\Sigma$ is a run in $\Sigma^\prime$. 
\end{proof}

We denote that a $K$-bounded coarse run \gls*{run_coarse} satisfies an \gls*{stl} formula \gls*{stlformula} by $\gls*{run_coarse} \gls*{sat}_0^K \gls*{stlformula}$ and write $\gls*{run_coarse} \gls*{sat}_k^K \gls*{stlformula}$ when a coarse run satisfies the \gls*{stl} formula at instant $t_k$. Note that a necessary and sufficient condition for a transition $(\gls*{poly}_k,\gls*{poly}_{k+1}) \in \delta^\prime$ is that there exists a predicate $\idstlpred$ such that $\idstlpred \in \mathcal{L}(\gls*{poly}_k)$ and $\idstlpred \in \mathcal{L}(\gls*{poly}_{k+1})$, since $\gls*{stlfunc}(\gls*{run})>0$ is a convex polyhedron $\gls*{poly} \in \mathcal{S}^\prime$. Therefore, we define the coarse \gls*{stl} semantics as \gls*{stl} semantics except for the predicate, i.e., $\gls*{run_coarse} \gls*{sat}_k^K \idstlpred$ if and only if $\gls*{stlfunc}(\gls*{run_coarse}_k) > 0$ and $\gls*{stlfunc}(\gls*{run_coarse}_{k+1}) > 0$.   

\begin{example}\label{ex:noloop}
Let us consider a system $\gls*{statec}_{k+1} = \gls*{statec}_k + \gls*{ts} \gls*{inputc}_k$, where $\gls*{statec}_k = [x,y]^\intercal$ and $\boldsymbol{u}_k = [v_x,v_y]^\intercal$, starting $\gls*{statecini} = [5,5]^\intercal$ with sampling time $\gls*{ts} = 1$s which must satisfy a reach-avoid specification, i.e., $\gls*{stlformula} = \idstluntil{\gls*{stlformula}_{safe}}{[10,60]}{\gls*{stlformula}_{goal}}$, where $\gls*{stlformula}_{safe} = (x<10 \gls*{or} x>20 \gls*{or} y<10 \gls*{or} y>20) \gls*{and} (0<x<30) \gls*{and} (0<y<30)$, and $\gls*{stlformula}_{goal} = 20<x<30 \gls*{and} 0<y<10$. A run \gls*{run_coarse} with $K=10$ that satisfies \gls*{stlformula} is $\gls*{run_coarse} = ([5,5]^\intercal,[0,0]^\intercal)([25,5]^\intercal,[0,0]^\intercal)$ $\dots([25,5]^\intercal,[0,0]^\intercal)$. 
 
\end{example}

Now we can formally define the discrete planning problem as follows:

\begin{problem}\label{prob:dplan}
Given a time index length $K$, an \gls*{stl} formula \gls*{stlformula}, an initial state \gls*{statecini}, a sampling time $\gls*{ts}$, and a finite set of counterexamples $\gls*{polyseq}_{cex} = \{ \gls*{polyseq}^{[1]}_{cex},\gls*{polyseq}^{[2]}_{cex},\dots,\gls*{polyseq}^{[N_{cex}]}_{cex} \}$, a discrete planning problem is a feasibility problem as follows,
\begin{align}
\text{find } & \gls*{polyseq} \nonumber \\
\text{ s.t. } & \gls*{polyseq} = \gls*{abs}(\gls*{run_coarse}_0)\gls*{abs}(\gls*{run_coarse}_1)\dots\gls*{abs}(\gls*{run_coarse}_K), \label{eq:dplan_abs} \\
& \gls*{statec}_0 = \gls*{statecini} \text{ and } \gls*{run_coarse} \gls*{sat}_0^K \gls*{stlformula} \label{eq:dplan_stl} \\
& \forall i \in [1..N_{cex}] \exists k \in [0,k^{\prime[i]}] \gls*{run_coarse}_k \not\in \gls*{poly}_{cex,k}^{[i]}. \label{eq:dplan_cex}
\end{align}
\end{problem}
The constraint (\ref{eq:dplan_stl}) ensures adjacency and initial conditions $\gls*{statec}_0=\gls*{statecini}$. Constraint (\ref{eq:dplan_cex}) states that no solution can have same prefix as the counterexamples. Finally, constraint (\ref{eq:dplan_abs}) abstracts the coarse run into a discrete plan \gls*{polyseq}.  

\subsection{Constraint Satisfaction Problem}

Formally, a \gls*{csp} is defined as a triple \gls*{cspprob}, where $\gls*{cspvarset} = \{ \gls*{cspvar}_1,\gls*{cspvar}_2,\dots,\gls*{cspvar}_N \}$ is a finite set of variables, $\gls*{cspdomset} = \{ \gls*{cspdom}_1,\gls*{cspdom}_2,\dots,\gls*{cspdom}_N \}$ is a set of respective domains of values, and $\gls*{cspconsset} = \{ \gls*{cspcons}_1,\gls*{cspcons}_2,\dots,\gls*{cspcons}_M \}$ is a set of constraints \cite{kira:Russell:2009}. A variable $\gls*{cspvar}_i$ can take values on the nonempty domain $\gls*{cspdom}_i$. A constraint $\gls*{cspcons}_j$ is a pair $\langle t_j, R_j \rangle$, where $t_j \subseteq V$ is a subset of $n$ variables and $R_j$ is a $n$-ary relation on the corresponding subset of domains $\gls*{cspdom}_j$. An evaluation $\boldsymbol{v} : V \mapsto D$ of the variables $V$ is a function that assign a particular set of values in the corresponding subset of domains. We denote $\boldsymbol{v} \gls*{sat} \gls*{cspcons}_j$ if the values assigned to the variables $t_j$ satisfies the relation $R_j$, and $\boldsymbol{v} \gls*{sat} C$ if for all $j \in [1..M]$, $\boldsymbol{v} \gls*{sat} \gls*{cspcons}_j$. Note that we use the following truth function symbols for expressing logic relation $R_j$ in first order logic formulas: $\gls*{pred} | \gls*{true} | \gls*{false} | \gls*{neg} | \gls*{and} | \gls*{or} | \gls*{implies} | \gls*{iff}$, where $\gls*{pred} = \gls*{propvar} | R$ is an atomic proposition, $\gls*{propvar}$ is a propositinal variable, $R$ is a linear equality or inequality over integer or real variables, $\gls*{and}$ is conjunction, $\gls*{or}$ is disjunction, $\gls*{neg}$ is negation, $\gls*{true} = \gls*{pred} \gls*{or} \gls*{neg} \gls*{pred}$, $\gls*{false} = \gls*{pred} \gls*{and} \gls*{neg} \gls*{pred}$, $\gls*{pred}_1 \gls*{implies} \gls*{pred}_2 = \gls*{neg} \gls*{pred}_1 \gls*{or} \gls*{pred}_2$, and $\gls*{pred}_1 \gls*{iff} \gls*{pred}_2 = (\gls*{pred}_1 \gls*{implies} \gls*{pred}_2) \gls*{and} (\gls*{pred}_2 \gls*{implies} \gls*{pred}_1)$.

\begin{example}
Robot position $[x,y]^\intercal \in \mathbb{R}^2$ and the loop length $L \in \mathbb{N}$ are example of variables with respective domains. A constraint can be defined as a first order logic formula $L=1 \gls*{implies} x = 0$, meaning that $L=1$ only if $x=0$. Formally, this constraint is represented as a pair $\langle t, R \rangle$, where $t := \{ x, L \}$ and $R := L=1 \gls*{implies} x = 0$.
 
\end{example}

It is easily seen that a \gls*{csp} encoding will generate only finite length runs. However, Problem \ref{prob:dplan} is an existential model checking problem, and the abstracted system \gls*{tsys}$^\prime$ is a finite transition system. Hence, we can represent infinite runs that satify an \gls*{stl} formula in $(K,L)-$loop form \cite{Biere1999,schuppan2006linear}.

\begin{definition}
A run $\gls*{run}$ is a run in $(K,L)-$loop form if and only if $\gls*{run} = \gls*{run}_0\gls*{run}_1\dots\gls*{run}_{L-1}(\gls*{run}_L\dots\gls*{run}_K)^\omega$, where $K$ is length of $\gls*{run}$, $\gls*{run}_0\gls*{run}_1\dots\gls*{run}_{L-1}$ is a prefix of length $L$, and $(\gls*{run}_L\dots\gls*{run}_K)^\omega$ is an infinite loop. There is a loop when $L \leq K$ and $\gls*{run}_K = \gls*{run}_{L-1}$. 
\end{definition}

\begin{example}\label{ex:loop}
The run in Example \ref{ex:noloop} can be compactly represented in the $(K,L)$-loop form: $\gls*{run_coarse} = ([5,5]^\intercal,[0,0]^\intercal)$ $([25,5]^\intercal,[0,0]^\intercal)\big(([25,5]^\intercal,$ $[0,0]^\intercal)\big)^\omega$ with $K=2$ and $L=2$.  
 
\end{example}

Note that, unlike \gls*{ltl}, \gls*{stl} expresses delays and deadlines. Thus, we cannot directly apply the results in \cite{Biere1999,schuppan2006linear}. Hence, we define an \gls*{stl} semantics such that we can check satisfaction of an infinite run using a bounded algorithm. 
\begin{definition}\label{def:coarsesemantics}
The validity of an \gls*{stl} formula for a coarse run \gls*{run_coarse} in $(K,L)$-loop form is defined as follows,
\begin{itemize}
  \item $\gls*{run_coarse} \gls*{sat}_k^{(K,L)} \idstlpred$ if and only if $\gls*{stlfunc}(\gls*{run_coarse}_k) > 0$ and $\gls*{stlfunc}(\gls*{run_coarse}_{k+1}) > 0$;
  \item $\gls*{run_coarse} \gls*{sat}_k^{(K,L)} \gls*{neg} \idstlpred$ if and only if $\gls*{run_coarse} \gls*{sat}_k^{(K,L)} \idstlpred[-\gls*{stlfunc}]$;
  \item $\gls*{run_coarse} \gls*{sat}_k^{(K,L)} \gls*{stlformula}_1 \gls*{and} \gls*{stlformula}_2$ if and only if $\gls*{run_coarse} \gls*{sat}_k^{(K,L)}{\gls*{stlformula}_1}$ and $\gls*{run_coarse} \gls*{sat}_k^{(K,L)}{\gls*{stlformula}_2}$;
  \item $\gls*{run_coarse} \gls*{sat}_k^{(K,L)} \gls*{stlformula}_1 \gls*{or} \gls*{stlformula}_2$ if and only if $\gls*{run_coarse} \gls*{sat}_k^{(K,L)}{\gls*{stlformula}_1}$ or $\gls*{run_coarse} \gls*{sat}_k^{(K,L)}{\gls*{stlformula}_2}$;
  \item $\gls*{run_coarse} \gls*{sat}_k^{(K,L)} \idstlalways{[a,b]}{\gls*{stlformula}}$ if and only if $\forall t_{k^\prime} \in I_1 \, \gls*{run_coarse} \gls*{sat}_{k^\prime}^{(K,L)}\gls*{stlformula}$, and $t_K < b+t_k$ only if $t_L \leq t_K$;
  \item $\gls*{run_coarse} \gls*{sat}_k^{(K,L)} \idstlevent{[a,b]}{\gls*{stlformula}}$ if and only if $\exists t_{k^\prime} \in I_1 \, \gls*{run_coarse} \gls*{sat}_{k^\prime}^{(K,L)}\gls*{stlformula}$, and $t_K < a+t_k$ only if $t_L \leq t_K$ and $t_K - t_L \leq b - a$;
  \item $\gls*{run_coarse} \gls*{sat}_k^{(K,L)} \idstluntil{\gls*{stlformula}_1}{[a,b]}{\gls*{stlformula}_2}$ if and only if $\exists t_{k^\prime} \in I_1$ s.t. $\gls*{run_coarse}  \gls*{sat}_{k^\prime}^{(K,L)}\gls*{stlformula}_2$, $\gls*{run_coarse} \gls*{sat}_{k^{\prime\prime}}^{(K,L)}\gls*{stlformula}_1$ for any $t_{k^{\prime\prime}} \in I_2$, and $t_K < a+t_k$ only if $t_L \leq t_K$ and $t_K - t_L \leq b - a$,
\end{itemize} 
where,
\begin{align*}
I_1 = & [\min(a+t_k,t_L),\min(b+t_k,t_K)] \\
I_2 = & 
\begin{cases}
[t_k,t_{k^\prime}] & \text{if } t_{k^\prime} \geq a + t_k \\
[\min(t_k,t_L),t_K] & \text{otherwise},
\end{cases} 
\end{align*}
\end{definition}

The \gls*{stl} semantics for coarse run \gls*{run_coarse} in $(K,L)$-loop form is equivalent to \gls*{stl} semantics for standard coarse run $\gls*{run_coarse}^\prime$.
\begin{lemma}\label{lem:coarsesemantics}
A coarse run in $(K,L)$-loop form $\gls*{run_coarse}$ satisfies an \gls*{stl} formula, i.e., $\gls*{run_coarse} \gls*{sat}_0^{(K,L)} \gls*{stlformula}$, if and only if the coarse run $\gls*{run_coarse}^\prime$ also satisfies the specifications, i.e., $\gls*{run_coarse}^\prime \gls*{sat}_0^{K^\prime} \gls*{stlformula}$, where,
\begin{align*}
\gls*{run_coarse} = & \gls*{run_coarse}_0\gls*{run_coarse}_1\dots\gls*{run_coarse}_{L-1}(\gls*{run_coarse}_L\dots\gls*{run_coarse}_K)^\omega \\
 \gls*{run_coarse}^\prime = &  \gls*{run_coarse}_0\gls*{run_coarse}_1\dots\gls*{run_coarse}_{L-1}\gls*{run_coarse}_L\dots\gls*{run_coarse}_{K}\gls*{run_coarse}_L\dots\gls*{run_coarse}_{K}\dots\gls*{run_coarse}_{K^\prime}.
\end{align*}
\end{lemma}
 \begin{proof}
($\Leftarrow$) It follows from the fact that $\gls*{run_coarse}^\prime = \gls*{run_coarse}$ when $K^\prime = K$ and $L = \infty$. 

($\Rightarrow$) If $K=K^\prime$, the proposition trivially follows. Thus, let us assume that $K < K^\prime$. Hence, the instant that $\gls*{run_coarse}^\prime_{k^\prime} = \gls*{run_coarse}_k$ is $t_k = \min(t_{k^\prime},(t_{k^\prime}-t_L)mod(t_K-t_L) + t_L)$, because if $t_{k^\prime} < t_L$, $(t_{k^\prime}-t_L)mod(t_K-t_L) = 0$; thus, $t_k = t_{k^\prime}$. First, if $\forall t_k \in I_1$, then $\gls*{run_coarse}_{k} = \gls*{run_coarse}_{k^\prime}^\prime$ for any $t_{k^\prime} \in [\min(a+t_k,t_L),\infty]$, and $[a+t_k,b+t_k] \subseteq [\min(a+t_k,t_L),\infty]$. Second, if $\exists t_k \in I_1$ such that $\gls*{run_coarse}_{k} = \gls*{run_coarse}_{k^\prime}^\prime$, then $t_{k^\prime} \in [\min(a+t_k,t_L),\infty]$. If $a+t_k \leq t_K$, $t_{k^\prime} \in [a+t_k,b+t_k]$. Otherwise, if exists a loop ($t_L \leq t_K$) and the loop is shorter than $[a+t_k,b+t_k]$ ($t_K - t_L \leq b - a$), then $t_{k^\prime} \in [a+t_k,b+t_k]$. Finally, consider a third instant $t_{k^\prime} \in I_1$. If $t_{k^\prime} < a + t_k$, then $a + t_k > t_L$, $b + t_k > t_K$ and $t_{k^\prime} > t_L$. Hence, $t_{k^{\prime\prime}} \in [\min(t_k,t_L),t_K]$ implies that $t_{k^{\prime\prime}}^\prime \in [t_k,t_{k^\prime}^\prime]$ such that $t_{k^\prime}^\prime \in [a+t_k,b+t_k]$.  If $t_{k^\prime} \geq a + t_k$, then $t_{k^{\prime\prime}} \in [t_k,t_{k^\prime}]$ implies that $t_{k^{\prime\prime}}^\prime \in [t_k,t_{k^\prime}^\prime]$ such that $t_{k^\prime}^\prime \in [a+t_k,b+t_k]$. Therefore, $\gls*{run_coarse} \gls*{sat}_0^{(K,L)} \gls*{stlformula}$ implies that $\gls*{run_coarse}^\prime \gls*{sat}_0^{K^\prime} \gls*{stlformula}$.
\end{proof}
 
\begin{example}
Note that the $(K,L)$-loop run in Example \ref{ex:loop} satisfies $\idstluntil{\gls*{stlformula}_{safe}}{[10,60]}{\gls*{stlformula}_{goal}}$. Since $\gls*{run_coarse} \gls*{sat}_k^{(2,2)} \gls*{stlformula}_{safe}$ for any $k \in [0..2]$, $\gls*{run_coarse}_K = \gls*{run_coarse}_{L-1}$, $t_{k^\prime} = 2$s $\in [0,2] = I_1$, and $\gls*{run_coarse} \gls*{sat}_{k^\prime}^{(2,2)} \gls*{stlformula}_{goal}$, then $\gls*{run_coarse} \gls*{sat}_k^{10} \gls*{stlformula}_{safe}$ for any $k \in [0..10]$ and $\gls*{run_coarse} \gls*{sat}_{10}^{10} \gls*{stlformula}_{goal}$.   

\end{example}

Now, we formally define a \gls*{csp} that solves Problem \ref{prob:dplan}.
\begin{definition}\label{def:enc}
Given a length $K$, an \gls*{stl} formula \gls*{stlformula}, an initial state \gls*{statecini}, and a finite set of counterexamples $\gls*{polyseq}_{cex} = \{ \gls*{polyseq}^{[1]}_{cex},\gls*{polyseq}^{[2]}_{cex},\dots,\gls*{polyseq}^{[N_{cex}]}_{cex} \}$, a discrete planning encoding of an arbitrary \gls*{stl} formula \gls*{stlformula} is a \gls*{csp}, where,
\begin{itemize}
  \item $V = \{\gls*{run_coarse}_0,\gls*{run_coarse}_1,\dots,\gls*{run_coarse}_{K+1},L,\gls*{predset} \}$,
  \begin{itemize}
  \item $\gls*{run_coarse}_k = [\gls*{statec}_k^\intercal,\gls*{inputc}_k^\intercal]^\intercal$ is an evaluation of the coarse run $\gls*{run_coarse}$ at instant $k$,
  \item $L$ is an integer variable to assign a $(K,L)$-loop,
  \item $\gls*{predset}$ is set of $N_{\gls*{predset}}$ auxiliary propositional variables $\gls*{propvar}_{\gls*{stlformula}_i}^{k^\prime,k}$,$\gls*{stlformula}^\prime = [0..K]$, $k = [0..K]$, and each subformula $\gls*{stlformula}_i$ of \gls*{stlformula}, including the main formula,
  \end{itemize}
  \item $D = \{\mathbb{R}^{\gls*{statecnb}+\gls*{inputcnb}},\mathbb{R}^{\gls*{statecnb}+\gls*{inputcnb}},\dots,\mathbb{R}^{\gls*{statecnb}+\gls*{inputcnb}},\mathbb{N},\{ \gls*{true},\gls*{false} \}^{N_{\gls*{predset}}}  \}$,
  \item $C = \{ \gls*{cspcons}_{0,dom},\gls*{cspcons}_{1,dom},\dots,\gls*{cspcons}_{K,dom},\gls*{cspcons}_{\gls*{stlformula}},\gls*{cspcons}_{K,\gls*{stlformula}}$ is a set of constraints defined as follows,
  \begin{itemize}
    \item $\gls*{cspcons}_{k^\prime,dom} = \bigwedge_{k=0}^{k^\prime} \encformc{\gls*{stlformula}}_{k^\prime}^{k}$, where $\bigwedge_{k=0}^{k^\prime} \encformc{\gls*{stlformula}}_{k^\prime}^{k}$ constrains recursively variables $\gls*{propvar}_{\gls*{stlformula}_i}^{{k^\prime},k}$ such that $\gls*{propvar}_{\gls*{stlformula}_i}^{k,k}$ holds true only if $\gls*{run_coarse} \gls*{sat}_{k} \gls*{stlformula}_i$, 
    \item $\encformc{\gls*{stlformula}}_{k^\prime}^{k}$ is a recursive quantifier-free first order logic formula defined later,
    \item $\gls*{cspcons}_{\gls*{stlformula}} = \gls*{propvar}_{\gls*{stlformula}}^{0,0} \gls*{and} L > 0 \gls*{and} \gls*{statec}_0 = \gls*{statecini}$ states that $\gls*{run_coarse} \gls*{sat} \gls*{stlformula}$,
    \item $\gls*{cspcons}_{K,loop} = \bigwedge_{{k^\prime}=1}^K \Big( \big( L = {k^\prime} \gls*{implies} (\gls*{run_coarse}_K = \gls*{run_coarse}_{k^\prime-1} \gls*{and} \gls*{run_coarse}_{K+1} = \gls*{run_coarse}_{k^\prime}) \big) \gls*{and} \bigwedge_{k=0}^K  \encformb{\gls*{stlformula}}_K^{k}$ means that if a loop starts at an instant ${k^\prime}$, then the last state must be equal to $L-1$, and $\encformb{\gls*{stlformula}}_K^{k}$ defines recursively the loop constraints for the formula \gls*{stlformula},
    \item $\encformb{\gls*{stlformula}}_K^{k}$ is a recursive quantifier-free first order logic formula defined later,
    \item $\gls*{cspcons}_{cex} = \bigwedge_{i=1}^{N_{cex}} \bigvee_{k=1}^{k^{\prime[i]}} \gls*{run_coarse}_k \not\in \gls*{poly}_{cex,k}^{[i]}$ discards runs with prefixes like those in the conterexamples. 
  \end{itemize}
\end{itemize} 
\end{definition}
Note that a coarse run $\gls*{run_coarse}$ needs $K+1$ evaluations to generate a run of length $K$ because we have modified semantics for predicates, i.e., $\gls*{run_coarse} \gls*{sat}_k \idstlpred$ if and only if $\gls*{stlfunc}(\gls*{run_coarse}_k) > 0 \gls*{and} \gls*{stlfunc}(\gls*{run_coarse}_{k+1}) > 0$. We also need to constrain $\gls*{run_coarse}_K = \gls*{run_coarse}_{L-1} \gls*{and} \gls*{run_coarse}_{K+1} = \gls*{run_coarse}_L$ instead of just $\gls*{run_coarse}_K = \gls*{run_coarse}_{L-1}$ when a $(K,L)$-loop exists for the same reason. 

Since the semantics of an \gls*{stl} formula \gls*{stlformula} are defined recursively, we define recursive quantifier-free first order formulas $\encformc{\gls*{stlformula}}_{k^\prime}^{k}$ to evaluate the truth of an \gls*{stl} formula, and $\encformb{\gls*{stlformula}}_K^{k}$ to evaluate when a $(K,L)$-loop is necessary.
\begin{definition}\label{def:encforma}
The formula $\encformc{\gls*{stlformula}}_{k^\prime}^{k}$ specifies recursively that Boolean variable $\gls*{propvar}_{\gls*{stlformula}}^{k,k}$ holds true only if $\gls*{run_coarse} \gls*{sat}_{k}^{(K,L)} \gls*{stlformula}$,\\
$\encformc{\idstlpred}_{k^\prime}^{k} := \Big(\gls*{propvar}_{\idstlpred}^{{k^\prime},k} \gls*{implies} \encforma{\idstlpred}_{k^\prime}^{k}\Big),$\\
$\encformc{\gls*{neg} \idstlpred}_{k^\prime}^{k}  := \Big(\gls*{propvar}_{\gls*{neg} \idstlpred}^{{k^\prime},k} \gls*{implies} \encforma{\idstlpred[-\mu]}_{k^\prime}^{k}\Big),$\\
$\encformc{\gls*{stlformula}_1 \gls*{and} \gls*{stlformula}_2}_{k^\prime}^{k} := \Big( \gls*{propvar}_{\gls*{stlformula}_1 \gls*{and} \gls*{stlformula}_2}^{{k^\prime},k} \gls*{implies} \gls*{propvar}_{\gls*{stlformula}_1}^{{k^\prime},k} \gls*{and} \gls*{propvar}_{\gls*{stlformula}_2}^{{k^\prime},k}\Big) \gls*{and} \encformc{\gls*{stlformula}_1}_{k^\prime}^{k} \gls*{and} \encformc{\gls*{stlformula}_2}_{k^\prime}^{k},$ \\
$\encformc{\gls*{stlformula}_1 \gls*{or} \gls*{stlformula}_2}_{k^\prime}^{k} := \Big( \gls*{propvar}_{\gls*{stlformula}_1 \gls*{or} \gls*{stlformula}_2}^{{k^\prime},k} \gls*{implies} \gls*{propvar}_{\gls*{stlformula}_1}^{{k^\prime},k} \gls*{or} \gls*{propvar}_{\gls*{stlformula}_2}^{{k^\prime},k}\Big) \gls*{and} \encformc{\gls*{stlformula}_1}_{k^\prime}^{k} \gls*{and} \encformc{\gls*{stlformula}_2}_{k^\prime}^{k},$ \\
$\encformc{\idstlalways{[a,b]}{\gls*{stlformula}}}_{k^\prime}^{k} := \Big(\gls*{propvar}_{\idstlalways{[a,b]}{\gls*{stlformula}}}^{{k^\prime},k} \gls*{implies} \encforma{\idstlalways{[a,b]}{\gls*{stlformula}}}_{k^\prime}^{k}  \Big) \gls*{and} \encformc{\gls*{stlformula}}_{k^\prime}^{k},$ \\
$\encformc{\idstlevent{[a,b]}{\gls*{stlformula}}}_{k^\prime}^{k} := \Big(\gls*{propvar}_{\idstlevent{[a,b]}{\gls*{stlformula}}}^{{k^\prime},k} \gls*{implies} \encforma{\idstlevent{[a,b]}{\gls*{stlformula}}}_{k^\prime}^{k}  \Big) \gls*{and} \encformc{\gls*{stlformula}}_{k^\prime}^{k},$ \\
$\encformc{\idstluntil{\gls*{stlformula}_1}{[a,b]}{\gls*{stlformula}_2}}_{k^\prime}^{k} := \Big(\gls*{propvar}_{\idstluntil{\gls*{stlformula}_1}{[a,b]}{\gls*{stlformula}_2}}^{{k^\prime},k} \gls*{implies} \encforma{\idstluntil{\gls*{stlformula}_1}{[a,b]}{\gls*{stlformula}_2}}_{k^\prime}^{k}  \Big)$ \\ 
\hspace*{5cm} $\gls*{and}  \encformc{\gls*{stlformula}_1}_{k^\prime}^{k} \gls*{and} \encformc{\gls*{stlformula}_2}_{k^\prime}^{k},$\\
where, \\
$\encforma{\idstlpred}_{k^\prime}^{k} := {k^\prime}=k \gls*{and} \gls*{stlfunc}(\gls*{run_coarse}_k) > 0 \gls*{and} \gls*{stlfunc}(\gls*{run_coarse}_{k+1}) > 0,$ \\
$\encforma{\idstlalways{[a,b]}{\gls*{stlformula}}}_{k^\prime}^{k} := \hat{a}_k^L \leq {k^\prime} \leq \hat{b}_k \gls*{implies} \big(\gls*{propvar}_{\gls*{stlformula}}^{{k^\prime},{k^\prime}} \gls*{and} ({k^\prime} \geq \hat{b}_k \gls*{or} \gls*{propvar}_{\idstlalways{[a,b]}{\gls*{stlformula}}}^{{k^\prime}+1,k})\big),$ \\
$\encforma{\idstlevent{[a,b]}{\gls*{stlformula}}}_{k^\prime}^{k} := \hat{a}_k^L < {k^\prime} < \hat{b}_k \gls*{implies} \big(\gls*{propvar}_{\gls*{stlformula}}^{{k^\prime},{k^\prime}} \gls*{or} (k^\prime < \hat{b}_k \gls*{and} \gls*{propvar}_{\idstlevent{[a,b]}{\gls*{stlformula}}}^{{k^\prime}+1,k})\big),$ \\
$\encforma{\idstluntil{\gls*{stlformula}_1}{[a,b]}{\gls*{stlformula}_2}}_{k^\prime}^{k} := {k^\prime} \leq \hat{b}_k \gls*{implies} \Big((\gls*{propvar}_{\gls*{stlformula}_1}^{{k^\prime},{k^\prime}} \gls*{and} \gls*{propvar}_{\idstluntil{\gls*{stlformula}_1}{[a,b]}{\gls*{stlformula}_2}}^{{k^\prime}+1,k}) \gls*{or}  $ \\
\hspace*{1cm}  $\big(\hat{a}_k^L < k^\prime < \hat{b}_k \gls*{and} \gls*{propvar}_{\gls*{stlformula}_2}^{{k^\prime},{k^\prime}} \gls*{and} \gls*{propvar}_{\gls*{stlformula}_1}^{{k^\prime},{k^\prime}} \gls*{and} (k^\prime \leq \hat{a}_k \gls*{or} \gls*{propvar}_{\idstluntil{\gls*{stlformula}_1}{[a,b]}{\gls*{stlformula}_2}}^{{k^\prime}+1,k})\big),$\\
$\hat{a}_k = \lfloor a/\gls*{ts} \rfloor + k$, $\hat{b}_k = \lceil b/\gls*{ts} \rceil + k$, and $\hat{a}_k^L = \min(\hat{a}_k, L)$.
\end{definition}

\begin{example}\label{ex:csp1}
Let us consider a simple \gls*{stl} formula $\gls*{stlformula} = \idstlalways{[6,8]}{x>0}$ for a simple system $x = x + u$ starting at $x = -1$. Thus, $\gls*{cspcons}_{k^\prime,dom}$ constraints for $k=0,1$ and $2$ are, \\
$\gls*{cspcons}_{0,dom} =  (\gls*{propvar}_{\idstlalways{[6,8]}{x>0}}^{0,0} \gls*{implies} \gls*{propvar}_{\idstlalways{[6,8]}{x>0}}^{1,0}) \gls*{and} (\gls*{propvar}_{x>0}^{0,0} \gls*{implies} x_0 > 0 \gls*{and} x_1 > 0),$ \\ 
$\gls*{cspcons}_{1,dom} = \bigwedge_{k=0}^1 \Big( (\gls*{propvar}_{\idstlalways{[6,8]}{x>0}}^{1,k} \gls*{implies} \gls*{propvar}_{\idstlalways{[6,8]}{x>0}}^{1,k}) \gls*{and}$ \\
\hspace*{3cm} $(\gls*{propvar}_{x>0}^{1,k} \gls*{implies} k=1 \gls*{and} x_k > 0 \gls*{and} x_{k+1} > 0) \Big),$ \\
$\gls*{cspcons}_{2,dom} = \bigwedge_{k=0}^2 \Big( (\gls*{propvar}_{\idstlalways{[6,8]}{x>0}}^{2,k} \gls*{implies} \gls*{propvar}_{\idstlalways{[6,8]}{x>0}}^{2,k}) \gls*{and}$ \\
\hspace*{3cm} $(\gls*{propvar}_{x>0}^{2,k} \gls*{implies} k=2 \gls*{and} x_k > 0 \gls*{and} x_{k+1} > 0) \Big).$

\end{example}

Note that Def. \ref{def:encforma} does not define loop constraints defined in temporal operator semantics in Def. \ref{def:coarsesemantics}. As we will show later, we want to use the incremental \gls*{smt} to increase performance, and loop constraints depend on $K$ which changes at each iteration. Thus, we assert all constraints that depend on $K$ on the first-order logic formula $\encformb{\gls*{stlformula}}_K^{k}$.  
\begin{definition}\label{def:encforma2}
The formula $\encformb{\gls*{stlformula}}_K^{k}$ defines recursively when a loop is necessary for a formula \gls*{stlformula} as follows, \\
$\encformb{\idstlpred}_K^{k} = \encforma{\gls*{neg} \idstlpred}^{k} := \gls*{true},$ \\
$\encformb{\gls*{stlformula}_1 \gls*{and} \gls*{stlformula}_2}_K^{k} = \encformb{\gls*{stlformula}_1 \gls*{or} \gls*{stlformula}_2}_K^{k} := \encformb{\gls*{stlformula}_1}_K^{k} \gls*{and} \encformb{\gls*{stlformula}_2}_K^{k},$\\
$\encformb{\idstlalways{[a,b]}{\gls*{stlformula}}}_K^{k} := \Big(\big(\gls*{propvar}_{\idstlalways{[a,b]}{\gls*{stlformula}}}^{k,k} \gls*{and} K<\hat{b}_k\big) \gls*{implies} L \leq K\Big) \gls*{and} \encformb{\gls*{stlformula}}_K^{k},$ \\
$\encformb{\idstlevent{[a,b]}{\gls*{stlformula}}}_K^{k} := \Big(\big(\gls*{propvar}_{\idstlevent{[a,b]}{\gls*{stlformula}}}^{k,k} \gls*{and} K<\hat{a}_k\big) \gls*{implies}$ \\
\hspace*{1.25cm} $\big(L \leq K \gls*{and} K-L \leq \hat{b}_k-\hat{a}_k-2\big)\Big) \gls*{and} \encformb{\gls*{stlformula}}_K^{k} \gls*{and} \gls*{neg} \gls*{propvar}_{\idstlevent{[a,b]}{\gls*{stlformula}}}^{K,k},$ \\
$\encformb{\idstluntil{\gls*{stlformula}_1}{[a,b]}{\gls*{stlformula}_2}}^{k} :=  \encformb{\gls*{stlformula}_1}_K^{k} \gls*{and} \encformb{\gls*{stlformula}_2}_K^{k} \gls*{and} \gls*{neg} \gls*{propvar}_{\idstluntil{\gls*{stlformula}_1}{[a,b]}{\gls*{stlformula}_2}}^{K,k} \gls*{and} $ \\
\hspace*{0.25cm} $\Big(\big(\gls*{propvar}_{\idstluntil{\gls*{stlformula}_1}{[a,b]}{\gls*{stlformula}_2}}^{k,k} \gls*{and} K<\hat{a}_k\big) \gls*{implies}\big(L \leq K \gls*{and} K-L \leq \hat{b}_k-\hat{a}_k-2\big)\Big).$ 

\end{definition}
Note that $\gls*{propvar}_{\idstlevent{[a,b]}{\gls*{stlformula}}}^{K,k}$ (and $\gls*{propvar}_{\idstluntil{\gls*{stlformula}_1}{[a,b]}{\gls*{stlformula}_2}}^{K,k}$) must hold false to ensure that $\gls*{propvar}_{\gls*{stlformula}}^{{k^\prime},{k^\prime}}$ ($\gls*{propvar}_{\gls*{stlformula}_2}^{{k^\prime},{k^\prime}}$) holds true for $k^\prime \in [0,K]$.   

\begin{example}
Continuing from Example \ref{ex:csp1}, $\gls*{cspcons}_{K,loop}$ for $K=2$ is , \\
$\gls*{cspcons}_{2,loop} = \bigwedge_{k=0}^1 (\gls*{propvar}_{\idstlalways{[6,8]}{x>0}}^{k,k} \gls*{implies} L \leq 0) \gls*{and}$ \\
$\bigwedge_{k=1}^2 \big(L=k \gls*{implies} (x_{k-1} = x_2 \gls*{and} x_k = x_3 \gls*{and} u_{k-1} = u_2 \gls*{and} u_k = u_3) \big).$  
\end{example}

The \gls*{csp} of definition \ref{def:enc} solves Problem \ref{prob:dplan}.
\begin{lemma}\label{lem:dplan_sound}
Given a length $K$, an \gls*{stl} formula \gls*{stlformula}, an initial state \gls*{statecini}, a sampling time $\gls*{ts}$, and a finite set of counterexamples $\gls*{polyseq}_{cex} = \{ \gls*{polyseq}^{[1]}_{cex},\gls*{polyseq}^{[2]}_{cex},\dots,\gls*{polyseq}^{[N_{cex}]}_{cex} \}$, there exist an evaluation $\gls*{run_coarse}$ of the \gls*{csp} in Def. \ref{def:enc} if and only if the discrete plan $\gls*{polyseq} = \gls*{abs}(\gls*{run_coarse}_0)\gls*{abs}(\gls*{run_coarse}_1)\dots\gls*{abs}(\gls*{run_coarse}_{L-1})(\gls*{abs}(\gls*{run_coarse}_L)\dots\gls*{abs}(\gls*{run_coarse}_K))^\omega$ is a solution for Problem \ref{prob:dplan}.
\end{lemma}
\begin{proof}
($\Rightarrow$) Let $\gls*{run_coarse}$ be a envaluation of of the CSP in Def. \ref{def:enc}. First, we show that $\gls*{run_coarse} \gls*{sat}_0^{(K,L)} \gls*{stlformula}$ using the recursive properties of the STL semantics to prove by induction. From Lemma \ref{lem:coarsesemantics}, \gls*{run_coarse} generates a run $\gls*{run_coarse}^\prime$ which $\gls*{run_coarse}^\prime \gls*{sat}_k^{K^\prime} \gls*{stlformula}$. Next we prove that other constraints of Problem \ref{prob:dplan} are also satisfied.
 
First, note that, from $C_{k,dom}$, it follows that $k^\prime \geq k$. Moreover, $\gls*{propvar}_{\gls*{stlformula}}^{k,k}$ implies that $\encformc{\gls*{stlformula}}_{k}^{k}$,  hold true. Also, $t_{\hat{a}_k} \leq a + t_k$ and $t_{\hat{b}_k} \geq b + t_k$ implies that $ [t_{\hat{a}_k^L+1}..t_{\min(\hat{b}_k,K)-1}] \subseteq I_1$, $I_1 \subseteq [t_{\hat{a}_k^L}..t_{\min(\hat{b}_k,K)}]$, and $t_{\hat{b}_k}-t_{\hat{a}_k}-2 \leq b - a$ for any temporal operator. 
\textbf{predicate}: $\gls*{propvar}_{\idstlpred}^{k,k}$ implies that $\gls*{stlfunc}(\gls*{run_coarse}_k) > 0$ and $\gls*{stlfunc}(\gls*{run_coarse}_{k+1}) > 0$. Hence, $\gls*{run_coarse} \gls*{sat}_k^{(K,L)} \gls*{sat}_k^{(K,L)} \idstlpred$. 
\textbf{negation}: $\gls*{propvar}_{\gls*{neg} \idstlpred}^{k,k}$ implies that $-\gls*{stlfunc}(\gls*{run_coarse}_k) > 0$ and $-\gls*{stlfunc}(\gls*{run_coarse}_{k+1}) > 0$. Hence, $\gls*{run_coarse} \gls*{sat}_k^{(K,L)} \gls*{neg} \idstlpred$. 
\textbf{conjunction}: $\gls*{propvar}_{\gls*{stlformula}_1 \gls*{and} \gls*{stlformula}_2}^{k,k}$ and $\encformb{\gls*{stlformula}_1 \gls*{and} \gls*{stlformula}_2}_K^{k}$ implies that $\gls*{propvar}_{\gls*{stlformula}_1}^{k,k}$, $\gls*{propvar}_{\gls*{stlformula}_2}^{k,k}$, $\encformb{\gls*{stlformula}_1}_K^{k}$, and $\encformb{\gls*{stlformula}_2}_K^{k}$ hold true. Hence, $\gls*{run_coarse} \gls*{sat}_k^{(K,L)} \gls*{stlformula}_1 \gls*{and} \gls*{stlformula}_2$. 
\textbf{disjunction}: $\gls*{propvar}_{\gls*{stlformula}_1 \gls*{or} \gls*{stlformula}_2}^{k,k}$ and $\encformb{\gls*{stlformula}_1 \gls*{or} \gls*{stlformula}_2}_K^{k}$ implies that $\gls*{propvar}_{\gls*{stlformula}_1}^{k,k}$ or $\gls*{propvar}_{\gls*{stlformula}_2}^{k,k}$, $\encformb{\gls*{stlformula}_1}_K^{k}$, and $\encformb{\gls*{stlformula}_2}_K^{k}$ hold true. Hence, $\gls*{run_coarse} \gls*{sat}_k^{(K,L)} \gls*{stlformula}_1 \gls*{or} \gls*{stlformula}_2$. 
\textbf{always}: $\gls*{propvar}_{\idstlalways{[a,b]}{\gls*{stlformula}}}^{k,k}$ and $\encformb{\idstlalways{[a,b]}{\gls*{stlformula}}}_K^{k}$ implies that: (1) $\gls*{propvar}_{\gls*{stlformula}}^{k^\prime,k^\prime}$ holds true for $k^\prime \in [\hat{a}_k^L..\min(\hat{b}_k,K)]$, and (2) $K<\hat{b}_k$ only if $K\leq L$. Thus, $\gls*{run_coarse} \gls*{sat}_k^{(K,L)} \idstlalways{[a,b]}{\gls*{stlformula}}$.
\textbf{eventually}: $\gls*{propvar}_{\idstlevent{[a,b]}{\gls*{stlformula}}}^{k,k}$ and $\encformb{\idstlevent{[a,b]}{\gls*{stlformula}}}_K^{k}$ implies that: (1) there exists $k^\prime \in [\hat{a}_k^L+1..\min(\hat{b}_k,K)-1]$ such that $\gls*{propvar}_{\gls*{stlformula}}^{k^\prime,k^\prime}$ holds true, and (2) $K<\hat{a}_k$ only if $K\leq L$ and $K-L < \hat{b}_k-\hat{a}_k$. Hence, $\gls*{run_coarse} \gls*{sat}_k^{(K,L)} \idstlevent{[a,b]}{\gls*{stlformula}}$.
\textbf{until}: $\gls*{propvar}_{\idstluntil{\gls*{stlformula}_1}{[a,b]}{\gls*{stlformula}_2}}^{k,k}$ and $\encformb{\idstluntil{\gls*{stlformula}_1}{[a,b]}{\gls*{stlformula}_2}}^{k}$ implies that: (1) there exists $k^\prime \in [\hat{a}_k^L+1..\min(\hat{b}_k,K)-1]$ such that $\gls*{propvar}_{\gls*{stlformula}_2}^{k^\prime,k^\prime}$ holds true, (2) if $\hat{a}_k<k^\prime$ then $\gls*{propvar}_{\gls*{stlformula}_2}^{k^{\prime\prime},k^{\prime\prime}}$ for any $k^{\prime\prime} \in [k..\min(\hat{b}_k,K)]$, else $\gls*{propvar}_{\gls*{stlformula}_2}^{k^{\prime\prime},k^{\prime\prime}}$ for any $k^{\prime\prime} \in [k..k^\prime]$, and (3) $K<\hat{a}_k$ only if $K\leq L$ and $K-L < \hat{b}_k-\hat{a}_k$. Thus, $\gls*{run_coarse} \gls*{sat}_k^{(K,L)} \idstluntil{\gls*{stlformula}_1}{[a,b]}{\gls*{stlformula}_2}$. \textbf{any STL formula}: By induction, $\gls*{run_coarse} \gls*{sat}_k^{(K,L)} \gls*{stlformula}$.

Now, note that $C_{\gls*{stlformula}}$ ensures that $L$ is valid, i.e., $L > 0$, and $\gls*{statec}_0 = \gls*{statecini}$. Additionally, $C_{k,dom}$ holds true only if $\gls*{run_coarse}_k \in \mathbb{R}^{\gls*{statecnb}} \times \mathbb{R}^{\gls*{inputcnb}}$, and $C_{cex}$ only if $\forall i \in [1..N_{cex}] \exists k \in [0,k^{\prime[i]}] \gls*{run_coarse}_k \not\in \gls*{poly}_{cex,k}^{[i]}]$. Therefore, the proposition holds. 

($\Leftarrow$) We will prove by contradiction. Let $\gls*{polyseq}$ be a solution for Problem \ref{prob:dplan}; thus, $\gls*{run_coarse}$ is a coarse run which is not in $(K,L)$-loop form and $\gls*{polyseq} = \gls*{abs}(\gls*{run_coarse}_1)\gls*{abs}(\gls*{run_coarse}_1)\dots\gls*{abs}(\gls*{run_coarse}_K)$. First, we show that $\gls*{run_coarse}$ satisfy the constraints $C_{0,dom},C_{1,dom},\dots,C_{K,dom}$ using recursive properties of $\encformc{\gls*{stlformula}}_{k}^{k_0}$. Next, we show that this run also satisfy the other constraints in CSP. Thus, by contradiction the Proposition holds. 
 
Note that $\hat{a}_k < L$ and $\hat{b}_k < K$ for any temporal operator; thus, $I_1 = [\hat{a}_k,\hat{b}_k]$ and $I_2 = [k,k^\prime]$. 
\textbf{predicate}: $\gls*{run_coarse} \gls*{sat}_k^K \idstlpred$ implies that $\gls*{stlfunc}(\gls*{run_coarse}_k) > 0$ and $\gls*{stlfunc}(\gls*{run_coarse}_{k+1}) > 0$. Hence, $\gls*{propvar}_{\idstlpred}^{k,k}$ can hold true. 
\textbf{negation}: $\gls*{run_coarse} \gls*{sat}_k^K \gls*{neg} \idstlpred$ implies that $-\gls*{stlfunc}(\gls*{run_coarse}_k) > 0$ and $-\gls*{stlfunc}(\gls*{run_coarse}_{k+1}) > 0$. Thus, $\gls*{propvar}_{\gls*{neg} \idstlpred}^{k,k}$ can hold true.
\textbf{conjunction}: $\gls*{run_coarse} \gls*{sat}_k^K \gls*{stlformula}_1 \gls*{and} \gls*{stlformula}_2$ implies that $\gls*{run_coarse} \gls*{sat}_k^K \gls*{stlformula}_1$ and $\gls*{run_coarse} \gls*{sat}_k^K \gls*{stlformula}_2$. Hence, $\gls*{propvar}_{\gls*{stlformula}_1}^{k,k}$ and $\gls*{propvar}_{\gls*{stlformula}_2}^{k,k}$ can hold true, implying that $\gls*{propvar}_{\gls*{stlformula}_1 \gls*{and} \gls*{stlformula}_2}^{k,k}$ can hold true. 
\textbf{disjunction}: $\gls*{run_coarse} \gls*{sat}_k^K \gls*{stlformula}_1 \gls*{or} \gls*{stlformula}_2$ implies that $\gls*{run_coarse} \gls*{sat}_k^K \gls*{stlformula}_1$ or $\gls*{run_coarse} \gls*{sat}_k^K \gls*{stlformula}_2$. Hence, $\gls*{propvar}_{\gls*{stlformula}_1}^{k,k}$ or $\gls*{propvar}_{\gls*{stlformula}_2}^{k,k}$ can hold true, implying that $\gls*{propvar}_{\gls*{stlformula}_1 \gls*{or} \gls*{stlformula}_2}^{k,k}$ can hold true.
\textbf{always}: If $\gls*{run_coarse} \gls*{sat}_k^K \idstlalways{[a,b]}{\gls*{stlformula}}$, then for all $t_k \in [a+t_k, b+t_k]$, $\gls*{run_coarse} \gls*{sat}_k^K \gls*{stlformula}$. Consequently, $\gls*{propvar}_{\gls*{stlformula}}^{k^\prime,k^\prime}$ can hold true for any $k^\prime \in [\hat{a}_k,\hat{b}_k]$, implying that $\gls*{propvar}_{\idstlalways{[a,b]}{\gls*{stlformula}}}^{k,k}$ can hold true. 
\textbf{eventually}: If $\gls*{run_coarse} \gls*{sat}_k^K \idstlevent{[a,b]}{\gls*{stlformula}}$, then there exists $t_k \in [a+t_k, b+t_k]$ such that $\gls*{run_coarse} \gls*{sat}_k^K \gls*{stlformula}$. Consequently, $k^\prime \in [\hat{a}_k,\hat{b}_k]$ and $\gls*{propvar}_{\gls*{stlformula}}^{k^\prime,k^\prime}$ can hold true, implying that $\gls*{propvar}_{\idstlevent{[a,b]}{\gls*{stlformula}}}^{k,k}$ can hold true. 
\textbf{until}: $\gls*{run_coarse} \gls*{sat}_k^K \idstluntil{\gls*{stlformula}_1}{[a,b]}{\gls*{stlformula}_2}$ implies that: (1) there exists $t_{k^\prime} \in [a+t_k, b+t_k]$ such that $\gls*{run_coarse} \gls*{sat}_k^K \gls*{stlformula}_2$, and (2) for any $t_{k^{\prime\prime}} \in [t_k,t_{k^\prime}]$ $\gls*{run_coarse} \gls*{sat}_k^K \gls*{stlformula}_1$. Thus, $k^\prime \geq \hat{a}_k$ and $t_{k^\prime} \in [\hat{a}_k, \hat{b}_k]$, implying that $\gls*{propvar}_{\idstlalways{[a,b]}{\gls*{stlformula}}}^{k,k}$ can hold true. 
\textbf{any STL formula}: By induction, if $\gls*{run_coarse} \gls*{sat}_k^K \gls*{stlformula}$, then $\gls*{propvar}_{\gls*{stlformula}}^{k,k}$ can hold true. 
\textbf{constraints in C}: Since $L > K \geq 0$ and $\gls*{statec}_0 = \gls*{statecini}$, $C_{\gls*{stlformula}}$ holds true. Thus, $C_{k,dom}$ holds true for any $k \in [0..K]$. Since $L>K$, $C_{K,loop}$ also holds true. Finally, $C_{cex}$ must hold true if $\forall i \in [1..N_{cex}] \exists k \in [0,k^{\prime[i]}] \gls*{run_coarse}_k \not\in \gls*{poly}_{cex,k}^{[i]}]$. Therefore, by contradiction, it follows that the Theorem holds.
\end{proof}

\begin{remark}
One of the distinct feature our proposed encoding is the separation of constraints that depend on the length $K$ ($\encformb{\gls*{stlformula}}_K^{k}$) to other constraints ($C_{k,dom}$), which changes during the search. Thus, we prepare this encoding to use with incremental \gls*{smt} solvers which can significantly improve scalability. To solve Problem \ref{prob:dplan}, we also include the counterexample constraints $C_{cex}$, which make sure that we avoid finding solution which are known to be dynamically unfeasible.  
\end{remark}

\subsection{Algorithm}

As shown in Fig. \ref{fig:diag1}, discrete plans are generated and evaluated using an iterative deepening search. This search iteratively increases the length ($K = K + 1$) starting from $K = 0$. Thus, as illustrated in Algorithm \ref{algo:dplan}, we leverage incremental solution capabilities of \gls*{smt} solvers such as Z3 \cite{de2008z3} to add and remove constraints as the search is deepened to use the lemmas found in previous searches, increasing the solver performance. We denote the assertion of a constraint in an \gls*{smt} solver by $SMT \gls*{assert} \gls*{cspcons}_i$. At each deepening iteration, we only add constraints for the current instant of time (line \ref{alg:dplan_add}). When we receive a new counterexample ($\gls*{polyseq}_{cex}\ \neq \gls*{emptyset}$), it means that a discrete plan was discarded by the continuous planning. It means that we did not deepen the search; thus, we only add these counterexamples encoding. Finally, if the encoding is satisfiable, we extract a run $\gls*{run_coarse}$ that is decoded into a discrete plan $\gls*{polyseq} = \gls*{abs}(\gls*{run_coarse}_0)\gls*{abs}(\gls*{run_coarse}_1)\dots\gls*{abs}(\gls*{run_coarse}_K)$, or return unsatisfiable.

\begin{algorithm}
    \caption{dplan}\label{algo:dplan}
    \begin{algorithmic}[1]
		\REQUIRE \idstldplaninput
		\ENSURE \idstldplanoutput
        \IF{$K = 0$}
            \STATE $SMT \gls*{assert} \gls*{cspcons}_{\gls*{stlformula}}$
        \ELSE
        	\STATE $SMT.pop$
        \ENDIF
        \STATE $SMT \gls*{assert} \begin{cases}
        \gls*{cspcons}_{cex} & \textrm{if } \gls*{polyseq}_{cex} \neq \gls*{emptyset} \\
        \gls*{cspcons}_{K,dom} & \text{otherwise}
        \end{cases}$ \label{alg:dplan_add}
        \STATE $SMT.push$
        \STATE $SMT \gls*{assert} \gls*{cspcons}_{K,loop}$
        \STATE $\langle \text{Status}, \gls*{run_coarse}, L \rangle \gls*{assert} SMT.\text{check}$
        \IF{$\text{Status} = sat$}
            \STATE $\gls*{polyseq} \gls*{assert} \gls*{abs}(\gls*{run_coarse}_0)\gls*{abs}(\gls*{run_coarse}_1)\dots\gls*{abs}(\gls*{run_coarse}_K)$
        \ENDIF
  \end{algorithmic}
\end{algorithm}

The discrete plans generated by Algorithm \ref{algo:dplan} are provably correct.
\begin{theorem}\label{teo:dplan_alg_sound}
Given a length $K$, an \gls*{stl} formula \gls*{stlformula}, a finite set of counterexamples $\gls*{polyseq}_{cex} = \{ \gls*{polyseq}^{[1]}_{cex},\gls*{polyseq}^{[2]}_{cex}\dots\gls*{polyseq}^{[N_{cex}]}_{cex} \}$, a sampling time $\gls*{ts}$, and an initial state \gls*{statecini}, Algorithm \ref{algo:dplan} returns a satisfying discrete plan $\gls*{polyseq} = \gls*{abs}(\gls*{run_coarse})$ if and only if this discrete plan is a solution for Problem \ref{prob:dplan}.
\end{theorem}
\begin{proof}
Since Algorithm \ref{algo:dplan} solves a \gls*{csp} with linear constraints by \gls*{stl} definition, there exists sound and complete \gls*{smt} solver for the \gls*{csp} in Def. \ref{def:enc} \cite{de2011satisfiability}. From Lemma \ref{lem:dplan_sound}, the Theorem holds.    
\end{proof}
\begin{remark}
The implementation of the \gls*{csp} is not only decidable (sound and complete) but also is a scalable implementation. First, it is known that solving quantifier free first order logic with linear constraints is polynomical time in the worst case \cite{de2011satisfiability}. Moreover, we implement a incremental implementation of the \gls*{csp} which allows us to keep lemmas from previous steps.     
\end{remark}

\section{Feasibility Search}\label{sec:idstl}

Our feasibility search finds a dynamically feasible run from a given discrete plan. Algorithm \ref{algo:feas} illustrate our approach. We combine best first search with binary search tree to search for all possible candidate prefixes. These prefixes are generated by the following function:
\begin{align*}
& prefix(\gls*{polyseq},\gls*{preidx},\gls*{tlenvec}) := \gls*{poly}_0 \big(convhull(\gls*{poly}_0\cup\gls*{poly}_1)\big)^{\gls*{tlen}_{1}} \gls*{poly}_1 \dots \hspace{1cm} \\
& \pushright{\big(convhull(\gls*{poly}_k\cup\gls*{poly}_{k+1})\big)^{\gls*{tlen}_{k+1}} \gls*{poly}_{k+1} \dots\gls*{poly}_{\gls*{preidx}}.}
\end{align*}
Observe that we never calculate the convex hull from $\gls*{poly}_k$ and $\gls*{poly}_{k+1}$, because this convex hull can be efficiently obtained from the coarse run $\gls*{run_coarse}$ as follows,
\begin{align*}
& convhull(\gls*{poly}_k\cup\gls*{poly}_{k+1}) := \{ \gls*{run}_k \in \mathbb{R}^{\gls*{statecnb}+\gls*{inputcnb}}: \hspace{3cm} \\
& \pushright{\bigwedge_{\forall \idstlpred \in \mathcal{L}(\gls*{run_coarse}_k)\cap\mathcal{L}(\gls*{run_coarse}_{k+1})} \gls*{stlfunc}(\gls*{run}_k) > 0 \}.}
\end{align*}  

Since the dynamical constraints (\ref{eq:system}) are ordered (time-indexed), an infeasible prefix $\gls*{polyseq}^\prime$ is an \textit{irreductive consistent set} \cite{chinneck1991locating}. This means that feasibility of sufixes will not alter the infeasibility of the prefix. Furthermore, the complexity increases with the size of time indexes; thus, we start the search with shortest prefix $\gls*{polyseq}^\prime =\gls*{poly}_0\gls*{poly}_1$ ($\gls*{preidx}=1$). Moreover, the time length $\gls*{tlen}_{\gls*{preidx}}$ to reach $\gls*{poly}_{\gls*{preidx}+1}$ from $\gls*{poly}_{\gls*{preidx}}$ is also sorted, because, once we find a length $\gls*{tlen}^*_{\gls*{preidx}}$ which there exists a dynamically feasible run, there will also exist a dynamically feasible run for any length $\gls*{tlen}_{\gls*{preidx}} > \gls*{tlen}^*_{\gls*{preidx}}$. Hence, we apply a binary search tree (lines \ref{algo:feas_bsearch1}-\ref{algo:feas_bsearch2}) to find the shortest time length $\gls*{tlen}^*_1$ to reach $\gls*{poly}_1$ from $\gls*{poly}_0$ (at initial state). Once we find a feasible run for this prefix, we increase the prefix index $\gls*{preidx}=\gls*{preidx}+1$ and also find the shortest time length $\gls*{tlen}^*_2$ to reach $\gls*{poly}_2$. Note, though, that prefixes ($\gls*{preidx}^\prime < \gls*{preidx}$) can affect feasibility of suffixes. Thus, if we do not find a dynamically feasible length for a segment $\gls*{preidx}$, increasing previous prefixes can affect the feasibility. Thus, we start increasing these prefixes (line \ref{algo:feas_prefix}). We repeat this procesure until we find a dynamically feasible run to $\gls*{poly}_2$, or we stop the algorithm and return counterexamples (lines \ref{algo:feas_infeas1} and \ref{algo:feas_infeas2}). Again we increase the prefix length until we find a dynamically feasible run that satisfies the specifications ($\idstlrobfunc[0]{\gls*{stlformula}} > 0$) or return counterexamples.

We propose a convex optimization problem to check if a prefix of the discrete plan $\gls*{polyseq}$ is dynamically feasible. The formal correctness of a run $\gls*{run_coarse}$ that validates a discrete plan is in terms of $\gls*{tolfeas}$-completeness \cite{gao2012delta}. 
\begin{lemma}\label{lem:cplan}
Given a prefix $\gls*{polyseq}^\prime = prefix(\gls*{polyseq},\gls*{preidx},\gls*{tlenvec})$ of a discrete plan $\gls*{polyseq}$ with length $K \geq \gls*{preidx}$ and a loop index $L$, a model (\ref{eq:system}), and an initial state $\gls*{statecini}$, there exists a dynamically feasible run \gls*{run} for the prefix $\gls*{polyseq}^\prime$ if and only if the solution to the following problem,
\begin{equation}\label{eq:contplan}
 \begin{array}{rll}
\min\limits_{\substack{
\gls*{inputc}_0,\dots,\gls*{inputc}_{K^\prime-1} \in \mathbb{R}^m \\ 
\gls*{slack}_1,\dots,\gls*{slack}_{K^\prime} \in \mathbb{R} \\
\gls*{statec}_0,\dots,\gls*{statec}_{K^\prime} \in \mathbb{R}^n \\ 
}} \quad & \sum_{k = 1}^{K^\prime} s_k &  \\
\text{s.t.} \quad & \gls*{statec}_0 = \gls*{statecini}, & \\
\forall k \in [0..K^\prime] & \gls*{run}_k \in \mathcal{P}_k,  \\
\forall k \in [0..K^\prime-1] & \| \gls*{statec}_{k+1} - A\gls*{statec}_k - B\gls*{inputc}_k \|_{\infty} \leq \gls*{slack}_{k+1}, \quad 0 \leq \gls*{slack}_{k+1}, \\
\forall k \in [1.K^\prime-1] & \gls*{slack}_k  \leq \gls*{slack}_{K^\prime}, \\
& \gls*{run}[N] = \gls*{run}_{L^\prime-1} \text{ only if } L \leq K \text{ and } \gls*{preidx} = K
  	\end{array}
\end{equation}
results into $\gls*{slack}_{K^\prime} \leq \gls*{tolfeas}$, where $K^\prime = K + \sum\limits_{k=1}^{K} \gls*{tlenvec}$ and $L^\prime = L +\sum\limits_{k=1}^{L-1} \gls*{tlenvec}$.
 \setlength{\arraycolsep}{5pt} 
\end{lemma}
\begin{proof}
First, note that the problem (\ref{eq:contplan}) is a \gls*{lp} problem; thus, there exists sound and complete solvers.
($\Leftarrow$) Assume that there exists a dynamically feasible run \gls*{run} for the prefix $\gls*{polyseq}^\prime$, but the solution to problem (\ref{eq:contplan}) is $\gls*{slack}_{K^\prime} > \gls*{tolfeas}$. Since there exist $\sum_{k = 0}^{\gls*{preidx}-1} s_k  \leq \gls*{tolfeas}$, $\gls*{slack}_{K^\prime}$ is not minimum. Thus, by contradiction, there exist a dynamically feasible run \gls*{run} for the prefix $\gls*{polyseq}^\prime$ only if the solution to (\ref{eq:contplan}) results into $\gls*{slack}_{K^\prime} \leq \gls*{tolfeas}$. ($\Rightarrow$) Now assume that solution for problem (\ref{eq:contplan}) results into $\gls*{slack}_{K^\prime} < \gls*{tolfeas}$. Thus, $\gls*{slack}_k < \gls*{tolfeas}$ for all $k \in [1..\gls*{preidx}]$, and the solution is dynamically feasible. Therefore, the proposition follows.   
\end{proof}

\begin{algorithm}[t]
	\caption{feas}\label{algo:feas}
    \begin{algorithmic}[1]
		\REQUIRE \idstlfeasinput
		\ENSURE \idstlfeasoutput
			\STATE $\gls*{tlenvec} \gls*{assert} \{ \gls*{tlen}_1 = 0$, $\dots$, $\gls*{tlen}_K = 0 \}$, $\quad \gls*{polyseq}_{cex} \gls*{assert} \emptyset$
			\FOR{\gls*{preidx} = 1 \TO K}
				\REPEAT
				\STATE $\gls*{tlen}_{\max} \gls*{assert}  \argmax\limits_{\gls*{tlen}_{\gls*{preidx}}} \idstlrobfunc{\gls*{stlformula}}(prefix(\gls*{polyseq},K,\gls*{tlenvec}),0) > 0$ 
				\STATE $\gls*{tlen}_{\min} \gls*{assert} -1$, $\quad \gls*{preidx}^\prime \gls*{assert} \gls*{preidx}-1$, $\quad \gls*{slack}_{K^\prime} \gls*{assert} \infty$
					\WHILE{$\gls*{tlen}_{\min} < \gls*{tlen}_{\max}$} \label{algo:feas_bsearch1}
						\STATE $\gls*{tlen}_N = \lceil \frac{\gls*{tlen}_{\max}+\gls*{tlen}_{\min}}{2} \rceil$
						\STATE $\langle \gls*{slack}_{K^\prime}^\prime,\gls*{run}^\prime \rangle \gls*{assert} \text{LP}(\gls*{polyseq},\gls*{preidx},\gls*{tlenvec},L, A, B,\gls*{statecini},\gls*{tolfeas})$ \label{algo:feas_lp}
						\IF{fval$^\prime\leq \gls*{tolfeas}$}
							\STATE $\gls*{tlen}_{\max} \gls*{assert} \gls*{tlen}_N - 1$, $\quad\gls*{slack}_{K^\prime} \gls*{assert} \gls*{slack}_{K^\prime}^\prime$, $\quad \gls*{run} \gls*{assert} \gls*{run}^\prime$
						\ELSE  
							\STATE $\gls*{tlen}_{\min} \gls*{assert} \gls*{tlen}_N$
						\ENDIF
					\ENDWHILE \label{algo:feas_bsearch2}
					\IF{fval$> \gls*{tolfeas}$} 
						\STATE $\gls*{tlen}_{\gls*{preidx}^\prime} \gls*{assert} \gls*{tlen}_{\gls*{preidx}^\prime} + 1$, $\quad \gls*{preidx}^\prime \gls*{assert} \begin{cases} 
						\gls*{preidx} - 1 & \text{if } \gls*{preidx}^\prime = 1 \\
						\gls*{preidx}^\prime - 1 & \text{otherwise}
						\end{cases}$ \label{algo:feas_prefix}
						\STATE $\gls*{polyseq}_{cex} \gls*{assert} \gls*{polyseq}_{cex} \cup \bigcup\limits_{\gls*{tlen}_{\gls*{preidx}} = [0..\gls*{tlen}_{\max}]} prefix(\gls*{polyseq},\gls*{preidx},\gls*{tlenvec})$ \label{algo:feas_infeas1}
					\ELSE
						\STATE $\gls*{tlen}_N \gls*{assert} \gls*{tlen}_{\max}$
						\STATE $\gls*{polyseq}_{cex} \gls*{assert} \gls*{polyseq}_{cex}  \cup \bigcup\limits_{\gls*{tlen}_{\gls*{preidx}} = [0..\gls*{tlen}_{\max}-1]} prefix(\gls*{polyseq},\gls*{preidx},\gls*{tlenvec})$ \label{algo:feas_infeas2}
					\ENDIF
				\UNTIL{fval$\leq \gls*{tolfeas}$ or $\gls*{tlen}_{\max} = 0$}
			\ENDFOR 
			\STATE isFeasible $\gls*{assert} \gls*{slack}_{K^\prime}\leq \gls*{tolfeas}$, $\quad L \gls*{assert} L + \sum\limits_{k=1}^{L-1} \gls*{tlenvec}$
\end{algorithmic}
\end{algorithm}

Therefore, Algorithm \ref{algo:feas} is sound and complete.
\begin{theorem}\label{theo:feas}
There exists a dynamically feasible run \gls*{run} for a discrete plan \gls*{polyseq} if and only if Algorithm \ref{algo:feas} finds a dynamically feasible run. 
\end{theorem}
\begin{proof}
($\Leftarrow$) Let $\gls*{run}$ be a dynamically feasible run such that $\gls*{run} \gls*{sat} \gls*{stlformula}$ and $\gls*{run} \in prefix(\gls*{polyseq},K,\gls*{tlenvec})$ for some $\gls*{tlenvec}$. Thus, $0 \leq \gls*{tlenvec} \leq \argmax\limits_{\gls*{tlen}_{\gls*{preidx}}} \idstlrobfunc{\gls*{stlformula}}(prefix(\gls*{polyseq},K,\gls*{tlenvec}),0) > 0$, meaning that Algorithm \ref{algo:feas} would solve problem (\ref{eq:contplan}) in line \ref{algo:feas_lp}. From Proposition \ref{prop:cplan}, it would mean that it would find the feasible run. Thus, by contradition,  There exists a dynamically feasible run \gls*{run} for a discrete plan \gls*{polyseq} only if Algorithm \ref{algo:feas} finds a dynamically feasible run.
 
($\Rightarrow$)  Let $\gls*{run}$ be a run found by Algorithm \ref{algo:feas} such that Status holds true. Thus, problem (\ref{eq:contplan}) in line \ref{algo:feas_lp} whould return $\gls*{slack}_{K^\prime} < \gls*{tolfeas}$, and, from Lemma \ref{lem:cplan}, it is a dynamically feasible run. Thereferore, the lemma holds.
\end{proof}

\begin{example}
Consider the discrete plan generated by the coarse $(K,L)$-loop run in Example \ref{ex:loop}. The resulting feasible trajectory is illustrated in Fig. \ref{fig:ex1_feas}. 
\begin{figure}
\centering
\begin{subfigure}[b]{0.2\textwidth}
\centering
\includegraphics[width=0.9\textwidth]{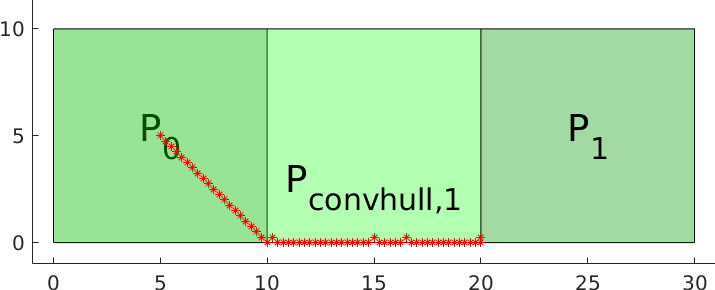}
\caption{Feasible}
\label{fig:ex1_feas}
\end{subfigure}
\begin{subfigure}[b]{0.20\textwidth}
\centering
\includegraphics[width=0.9\textwidth]{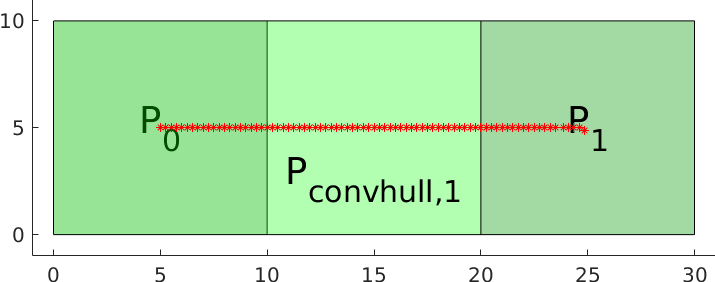}
\caption{Robust}
\label{fig:ex1_rob}
\end{subfigure}
\caption{Illustrative example of a robot trajectories in the configuration space.}
\label{fig:ex1_trajectories}
\end{figure}

\end{example}

\section{Robust Controller}\label{sec:robctrler}

We design a robust controller for the feasible plan in two steps. First, we increase the plan robustness using the robustness measure $\idstlrobfunc[0]{\gls*{stlformula}}$ of the STL quantitative semantics and generate a robust nominal run $\gls*{run}$. Second, we design a quadratic regulator to track this nominal run. 

We define a robust measure function $\gls*{dplanrobfunc}^{\gls*{polyseq}}_k$ for the discrete plan as follows,
\begin{equation}
\gls*{dplanrobfunc}^{\gls*{polyseq}}_k(\gls*{run}) = \min_{\forall\gls*{stlfunc} \in \gls*{poly}_k} \gls*{stlfunc}(\gls*{run}_k).  
\end{equation}  
Observe that \gls*{stlfunc} are facets of the convex constraint $\gls*{poly}_k$. We can use this robust measure to increase the \gls*{stl} robustness.  
\begin{theorem}
Given a dynamicall fesible run \gls*{run} that satisfies a discrete plan \gls*{polyseq}, i.e., $\gls*{polyseq}= \alpha(\gls*{run}_0)\dots\alpha(\gls*{run}_K)$, it holds that $\gls*{dplanrobfunc}^{\gls*{polyseq}}_0(\gls*{run}) \leq \idstlrobfunc[0]{\gls*{stlformula}}$.
\end{theorem}
\begin{proof}
From Lemma \ref{lem:dplan_sound}, $\gls*{run} \gls*{sat}_0^{(K,L)} \gls*{stlformula}$. 
\textbf{predicate}: If $\gls*{run} \gls*{sat}_k^{(K,L)} \idstlpred$, then, by Def. \ref{def:model}, $\idstlpred \in \mathcal{L}(\gls*{run}_k)$. Consequently, $\gls*{dplanrobfunc}^{\gls*{polyseq}}_k(\gls*{run}) \leq \idstlrobfunc[k]{\idstlpred}$. 
\textbf{negation}: Similarly, if $\gls*{run} \gls*{sat}_k^{(K,L)} \gls*{neg} \idstlpred$, then $\gls*{dplanrobfunc}^{\gls*{polyseq}}_k(\gls*{run}) \leq \idstlrobfunc[k]{\idstlpred[-\gls*{stlfunc}]}$. 
\textbf{conjunction}: If $\gls*{run} \gls*{sat}_k^{(K,L)} \gls*{stlformula}_1 \gls*{and} \gls*{stlformula}_2$, by Def. \ref{def:model}, $\idstlpred \in \mathcal{L}(\gls*{run}_k)$ for all predicates $\idstlpred$ in formulas $\gls*{stlformula}_1$ and $\gls*{stlformula}_2$. Thus, $\gls*{dplanrobfunc}^{\gls*{polyseq}}_k(\gls*{run}) \leq \min(\idstlrobfunc{\gls*{stlformula}_1}(\gls*{run},t_k),\idstlrobfunc{\gls*{stlformula}_2}(\gls*{run},t_k))$. 
\textbf{disjunction}: If $\gls*{run} \gls*{sat}_k^{(K,L)} \gls*{stlformula}_1 \gls*{or} \gls*{stlformula}_2$, by Def. \ref{def:model}, $\idstlpred \in \mathcal{L}(\gls*{run}_k)$ for all predicates $\idstlpred$ in the formula $\gls*{stlformula}_1$ or $\gls*{stlformula}_2$. Hence, $\gls*{dplanrobfunc}^{\gls*{polyseq}}_k(\gls*{run}) \leq \max(\idstlrobfunc{\gls*{stlformula}_1}(\gls*{run},t_k),\idstlrobfunc{\gls*{stlformula}_2}(\gls*{run},t_k))$. 
\textbf{always}: $\gls*{run} \gls*{sat}_k^{(K,L)} \idstlalways{[a,b]}{\gls*{stlformula}}$ implies that $\idstlpred \in \mathcal{L}(\gls*{run}_{k^\prime})$ for all predicates $\idstlpred$ in the formula $\gls*{stlformula}$ for any $k^\prime \in [a+t_k,b+t_k]$, by Def. \ref{def:model} and Lemma \ref{lem:coarsesemantics}. Hence, $\min_{k \in [k..K]}(\gls*{dplanrobfunc}^{\gls*{polyseq}}_k(\gls*{run})) \leq \min_{t_{k^\prime} \in [t_k+a,t_k+b]} \idstlrobfunc{\gls*{stlformula}}(\gls*{run},t_{k^\prime})$. 
\textbf{eventually}: $\gls*{run} \gls*{sat}_k^{(K,L)} \idstlevent{[a,b]}{\gls*{stlformula}}$ implies that there exists a $k^\prime \in [a+t_k,b+t_k]$ such that $\idstlpred \in \mathcal{L}(\gls*{run}_{k^\prime})$ for all predicates $\idstlpred$ in the formula $\gls*{stlformula}$, by Def. \ref{def:model} and Lemma \ref{lem:coarsesemantics}. Hence, $\min_{k \in [k..K]}(\gls*{dplanrobfunc}^{\gls*{polyseq}}_k(\gls*{run})) \leq \max_{t_{k^\prime} \in [t_k+a,t_k+b]} \idstlrobfunc{\gls*{stlformula}}(\gls*{run},t_{k^\prime})$. 
\textbf{until}: $\gls*{run} \gls*{sat}_k^{(K,L)} \idstluntil{\gls*{stlformula}_1}{[a,b]}{\gls*{stlformula}_2}$ implies that there exists a $k^\prime \in [a+t_k,b+t_k]$ such that $\idstlpred \in \mathcal{L}(\gls*{run}_{k^\prime})$ for all predicates $\idstlpred$ in the formula $\gls*{stlformula}_2$, and ${\idstlpred}^{\prime} \in \mathcal{L}(\gls*{run}_{k^{\prime\prime}})$ for any $k^{\prime\prime} \in [k..k^{\prime\prime}]$, by Def. \ref{def:model} and Lemma \ref{lem:coarsesemantics}. Hence, $\min_{k \in [k..K]}(\gls*{dplanrobfunc}^{\gls*{polyseq}}_k(\gls*{run})) \leq \max_{t_{k^\prime} \in [t_k+a,t_k+b]}\Big(\min\big( \idstlrobfunc{\gls*{stlformula}_2}(\gls*{run},t_{k^\prime}),\min_{t_{k^{\prime\prime}} \in [t_k,t_{k^\prime}]} \idstlrobfunc{\gls*{stlformula}_1}(\gls*{run},t_{k^{\prime\prime}})\big)\Big)$.
\textbf{\gls*{stl} formula}: By induction, $\gls*{dplanrobfunc}^{\gls*{polyseq}}_0(\gls*{run}) \leq \idstlrobfunc[0]{\gls*{stlformula}}$.
\end{proof} 
\begin{remark}
The simulation function ensures that if a predicate must holds, any run satifying the discrete plan will also satisfies the specifications. Thus, the robustness based on the discrete plan will always a lower bound for the \gls*{stl} robust semantics. 
\end{remark}

This robust measure function allows us define a \gls*{lp} problem that maximizes the robustnes for the discrete plan $\gls*{polyseq}^\prime = prefix(\gls*{polyseq},\gls*{preidx},\gls*{tlenvec})$.
\begin{equation}\label{eq:rplan}
 \begin{array}{rll}
\max\limits_{\substack{
\gls*{inputc}_0,\dots,\gls*{inputc}_{K^\prime-1} \in \mathbb{R}^{\gls*{inputcnb}} \\ 
\gls*{statec}_0,\dots,\gls*{statec}_{K^\prime} \in \mathbb{R}^{\gls*{statecnb}} \\ 
}} \quad & \min_{k \in [0..K^\prime]} \gls*{dplanrobfunc}^{\gls*{polyseq}}_k  &  \\
\text{s.t.} \quad & \gls*{statec}_0 = \gls*{statecini}, & \\
\forall k \in [1..K^\prime] & \sum_{k=0}^K \| \gls*{statec}_{k+1} - A\gls*{statec}_k - B\gls*{inputc}_k \|_{\infty} \leq \gls*{tolfeas}, \\
& \gls*{run}_{K} = \gls*{run}_{L^\prime-1} \text{ if } L \leq K \text{ and } N = K,
\end{array}
\end{equation}
where $K^\prime = K + \sum\limits_{k=1}^{K} \gls*{tlenvec}$ and $L^\prime = L +\sum\limits_{k=1}^{L-1} \gls*{tlenvec}$.
Note that this is a maxmin problem that can be encoded into a linear programming.   

Since dynamically feasible discrete plan found in Algorithm \ref{algo:feas} is minimum length, we propose a search algorithm that increases the time length of the plan to increase robutsness even further in Algorith \ref{algo:robctrler}. Note, though, we do not claim that there is not a longer time length where the robustness measure is greater. The main idea is to improve the robustness in a fast way. Thus, at each iteration, we increase the discrete plan and check if we can obtain a more robust run. If so, we keep this run. Otherwise, we discard. We repeat until we cannot improve the robustness. 

\begin{algorithm}
	\caption{rob}\label{algo:robctrler}
    \begin{algorithmic}[1]
		\REQUIRE \idstlrobinput
		\ENSURE \idstlroboutput
		\STATE $\langle \gls*{run}^\prime, \gls*{dplanrobfunc}^{\gls*{polyseq}}_k(\gls*{run})^\prime \rangle \gls*{assert} \text{LP}(\gls*{polyseq},K,\gls*{tlenvec},L,A,B,\gls*{statecini},\gls*{tolfeas})$
		\REPEAT
			\STATE $\gls*{run} \gls*{assert} \gls*{run}^\prime$, $\quad \gls*{dplanrobfunc}^{\gls*{polyseq}}_k(\gls*{run}) \gls*{assert} \gls*{dplanrobfunc}^{\gls*{polyseq}}_k(\gls*{run})^\prime$
			\FOR{\gls*{preidx} = 1 \TO K}
				\STATE $\gls*{tlenvec}^\prime \gls*{assert} \gls*{tlenvec}$, $\quad \gls*{tlen}_{\gls*{preidx}} \gls*{assert} \gls*{tlen}_{\gls*{preidx}}+1$
				\STATE $\langle \gls*{run}^\prime, \gls*{dplanrobfunc}^{\gls*{polyseq}}_k(\gls*{run})^\prime \rangle \gls*{assert} \text{LP}(\gls*{polyseq},K,\gls*{tlenvec}^\prime,L,A,B,\gls*{statecini},\gls*{tolfeas})$
				\IF{$\gls*{dplanrobfunc}^{\gls*{polyseq}}_k(\gls*{run})^\prime - \gls*{dplanrobfunc}^{\gls*{polyseq}}_k(\gls*{run}) \geq \gls*{tolfeas}$}
					\STATE $\gls*{run} \gls*{assert} \gls*{run}^\prime$, $\quad \gls*{tlen}_{\gls*{preidx}} \gls*{assert} \gls*{tlen}_{\gls*{preidx}}+1$
				\ENDIF
			\ENDFOR				
		\UNTIL{$\gls*{dplanrobfunc}^{\gls*{polyseq}}_k(\gls*{run})^\prime - \gls*{dplanrobfunc}^{\gls*{polyseq}}_k(\gls*{run}) < \gls*{tolfeas}$}
		\STATE $\gls*{fctrler} \gls*{assert} lqr(A,B,Q_f,Q,R)$
  \end{algorithmic}
\end{algorithm}

\begin{example}
Consider again the discrete plan generated by the coarse $(K,L)$-loop run in Example \ref{ex:loop}. We can see in Fig.~\ref{fig:ex1_feas} that this trajectory is not robust because is is close to boundaries defined in the predicates from specifications. However, Algorithm \ref{algo:robctrler} increases the robustness of this plan as shown in Fig.~\ref{fig:ex1_rob}. 
\end{example}

Now, we present the feedback controller design. Given a nominal run $\gls*{run}$, we stabilize the trajectory using finite-horizon, discrete-time LQR. First, we define a linear system around the nominal run as follows,
\begin{align}
\bar{\boldsymbol{x}}_k = & \gls*{statec}_k - \gls*{statec}_k^*, \quad \bar{\gls*{inputc}}_k = \gls*{inputc}_k - \gls*{inputc}_k^*, \\
\bar{\boldsymbol{x}}_{k+1} = & A \bar{\boldsymbol{x}}_k + B\bar{\gls*{inputc}}_k.
\end{align} 

Next, we define a quadratic regulator (tracking) cost function,
\begin{equation}
J = \bar{\boldsymbol{x}}_K^\intercal Q_f \bar{\boldsymbol{x}}_K + \sum_{k=0}^{K-1} \Big(\bar{\boldsymbol{x}}_k^\intercal Q \bar{\boldsymbol{x}}_k + \bar{\gls*{inputc}}_k^\intercal R \bar{\gls*{inputc}}_k  \Big),
\end{equation}
where $Q_f = Q_f^\intercal > 0$, $Q = Q^\intercal \geq 0$, and $R = R^\intercal > 0$.

Note that $Q$ and $R$ could easily be made functions of time as well. It can be shown that the optimal feedback controller is given by,
\begin{equation}
\gls*{inputc}_k = \gls*{inputc}_k^*-F_k (\gls*{statec}_k - \gls*{statec}_k^*)
\end{equation}
where,
\begin{equation}
F_k = (R + B^\intercal P_{k+1}B)^{-1}B^\intercal P_{k+1} A, 
\end{equation}
and $P_k$ is the solution to,
\begin{equation}
P_K = Q_f, \quad P_{k-1} = A^\intercal P_k A - A^\intercal P_k B(R + B^\intercal P_k B)^{-1}B^\intercal P_k A + Q.  
\end{equation}

\begin{example}
Let $Q_f = 10 I_2$, $Q = I_2$, and $R = I_2$, where $I_n$ is an identity matrix with dimension $n \times n$. Fig.~\ref{fig:ex1_result} illustrates $100$ executions, black dotted lines, for different initial states and a bounded disturbance in the model, i.e., $\gls*{statec}_{k+1} = \gls*{statec}_k + T_s \gls*{inputc}_k + \boldsymbol{w}$, where $\boldsymbol{w} \in [-0.1,0.1]^2$.  
\begin{figure}
\centering
\includegraphics[width=0.2\textwidth]{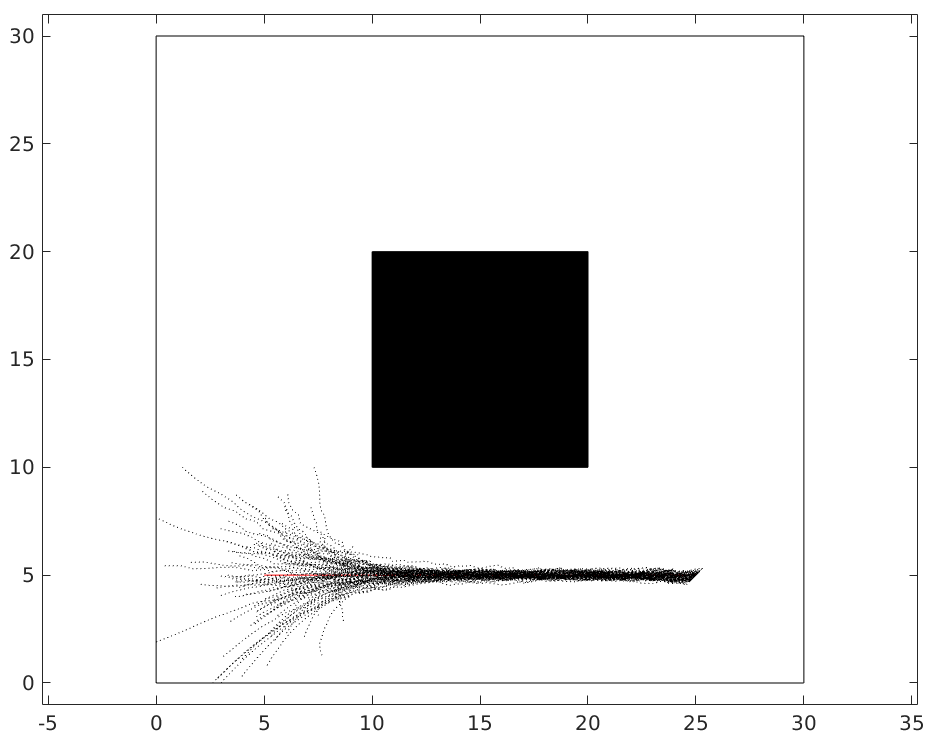}
\caption{Illustrative example of 100 robot trajectories including disturbances in the model and in the initial state.}
\label{fig:ex1_result}
\end{figure}
\end{example}

Now, we can show that our approach is sound and complete.
\begin{theorem}
Given a bounded time STL formula \gls*{stlformula}, a dynamic system (\ref{eq:system}) and an initial state $\gls*{statecini}$, then Algorithm \ref{algo:idstl} finds a reactive controller $\gls*{fctrler}$ if and only if the corresponding run $\gls*{run}$ satisfies the formula \gls*{stlformula}, i.e., $\gls*{run} \gls*{sat} \gls*{stlformula}$.
\end{theorem}
\begin{proof}
($\Leftarrow$) Assume that there exists a dynamically feasible run $\gls*{run}$ that satisfies the formula \gls*{stlformula}, but the  Algorithm \ref{algo:idstl} returns unsatisfiable. Hence, from Theorem \ref{teo:eabs}, there exists a discrete plan for the formula \gls*{stlformula}. Since there exists a finite bound \gls*{kmax} for a bounded time STL formula \gls*{stlformula} which is a conservative bound to ensure satisfiability and Algorithm \ref{algo:idstl} executes Algorithm \ref{algo:dplan} for $K\in[0..\gls*{kmax}]$, we conclude from Theorem \ref{teo:dplan_alg_sound} that Algorithm \ref{algo:idstl} finds a discrete plan for  \gls*{stlformula}. Moreover, Theorem \ref{theo:feas} implies that \ref{algo:idstl} finds dynamically feasible run that satisfies the formula \gls*{stlformula}. Since Algorithm \ref{algo:robctrler} always return a solution if there exists a dynamically feasible run, by contradiction, it follows that Algorithm \ref{algo:idstl} finds a reactive controller $\gls*{fctrler}$ that  the corresponding run $\gls*{run}$ that satisfies the formula \gls*{stlformula}. 

($\Rightarrow$) If Algorithm \ref{algo:idstl} finds a reactive controller $\gls*{fctrler}$, then, from Theorems \ref{teo:eabs}, \ref{teo:dplan_alg_sound} and \ref{theo:feas}, we conclude that the corresponding run $\gls*{run}$ that satisfies the formula \gls*{stlformula}. Therefore, the theorem follows.
\end{proof}

\section{Simulation}\label{sec:results}

In this section, we apply our framework to the problem of automated power line inspection with a quadrotor. 

\subsection{Quadrotor Dynamic Model}
The quadrotor is a well-modelled dynamical system in which four propellers are used to move a small aircraft. Assuming that damping and drag-like effects are negligible \cite{hehn2015real}, quadrotor dynamics have been shown to be differentially flat \cite{zhou2014vector}, meaning that states and inputs of the system can be written in terms of algebraic functions of appropriately chosen flat outputs and their derivatives. 

Following \cite{hehn2015real}, we assume that the yaw angle is always set to zero, and introduce a virtual input $\boldsymbol{v} \in \mathbb{R}^3$ that transforms the quadrotor dynamics into a linear chain of integrators
\begin{equation}
    \ddddot{\boldsymbol{p}} = \boldsymbol{v}
\end{equation}
where $\boldsymbol{p}=(x,y,z)$ is the quadrotor position in the world frame. 

\subsection{Power line Inspection Specification}

The task of power line inspection, which is typically performed by manually flying over a section of a power line, involves collecting images of key points for later evaluation.

\begin{figure}
	\begin{subfigure}{\linewidth}
		\centering
  		\includegraphics[width=0.7\linewidth]{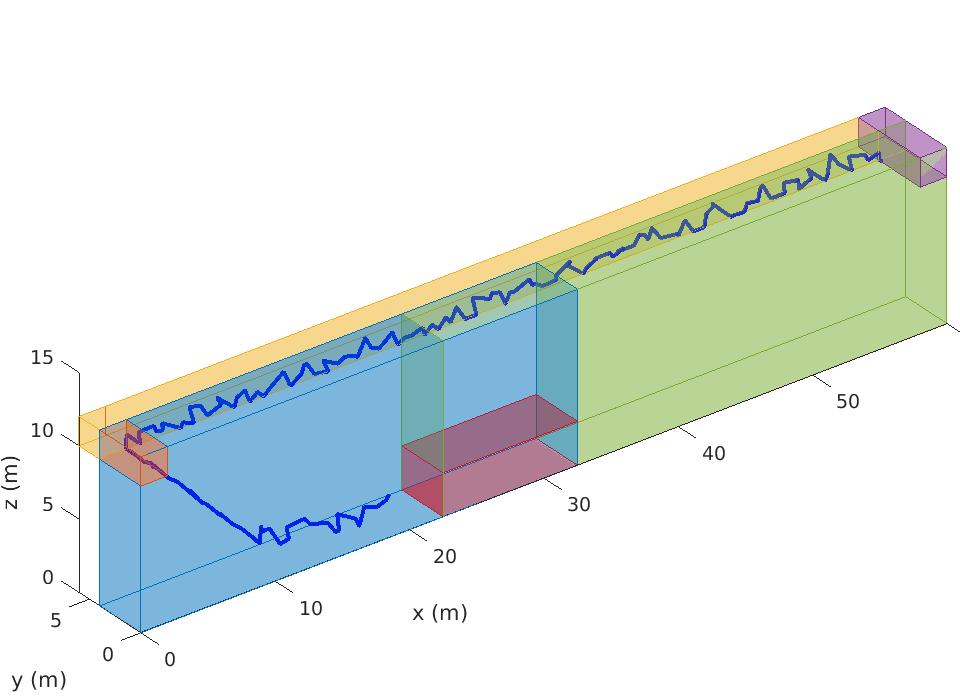}
    	\caption{Finding a dynamically feasible trajectory for a discrete plan. Starts with an initial condition at $(20,2,1.5)$, the discrete plan consist of the orange, yellow, purple, and red regions.The blue line is a feasible trajectory which reaches the purple discrete planning step.}
  		\label{fig:quadrrt}
	\end{subfigure}

	\begin{subfigure}{\linewidth}
		\centering
    	\includegraphics[width=0.85\linewidth]{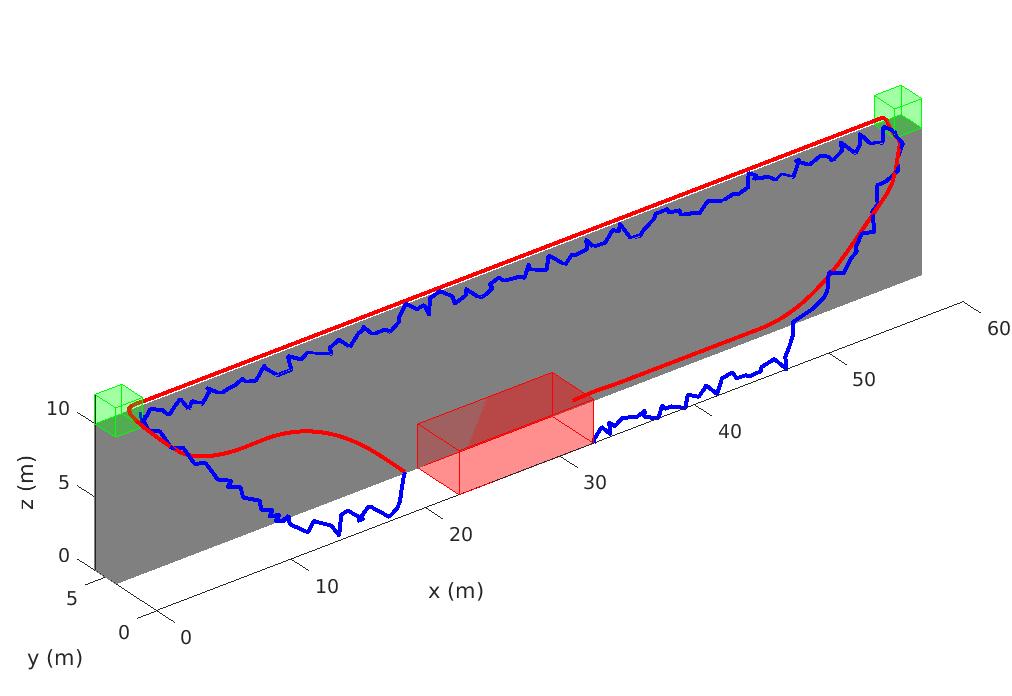}
   		\caption{A dynamically feasible (blue) and the final robust run (red).}
    	\label{fig:quad_robust}
	\end{subfigure}
    \caption{A quadrotor performing an infrastructure inspection task must visit two key points (green) and avoid a forbidden region (gray) before returning to a charging station (red). }
    \label{fig:quad_demo}
\end{figure}

A simple inspection task might be articulated as follows:
``Staying within the workspace area and away from the power line, take pictures of the top of two poles. Be sure to return to the charging station within 15 minutes to recharge the batteries.'' This simple specification, illustrated in Figure \ref{fig:quad_demo}, can be encoded in the following STL specification:

\begin{align*}
\gls*{stlformula} = & \idstlalways{[0,15min]}{\big(\gls*{stlformula}_{safe} \gls*{and} \gls*{stlformula}_W\big)} \gls*{and} \idstlevent{[0,15min]}{\Big((\gls*{stlformula}_{pole_1} \hspace{2cm} \\
& \pushright{\gls*{and} \idstluntil{\gls*{stlformula}_{Pline})}{[0,15min]}{\big(\gls*{stlformula}_{pole_2} \gls*{and} \idstlevent{[0,15min]}{\gls*{stlformula}_{home}} \big)\Big)}}},
\end{align*}
where,
\begin{itemize}
  \item $\gls*{stlformula}_{safe} = 4>y \gls*{or} y>6 \gls*{or} z>10$;
  \item $\gls*{stlformula}_W = 0<x<60 \gls*{and} 0<y<10 \gls*{and} 0<z<20$;
  \item $\gls*{stlformula}_{Pline} = 4<y<6 \gls*{and} 12>z$;  
  \item $\gls*{stlformula}_{pole_1} = 2>x \gls*{and} \gls*{stlformula}_{Pline}$;  
  \item $\gls*{stlformula}_{pole_2} = x>58 \gls*{and} \gls*{stlformula}_{Pline}$;
  \item $\gls*{stlformula}_{home} = 22.5<x<32.5 \gls*{and} 3>z$;  
\end{itemize}

Note that this specification is highly nonconvex, as it contains multiple nested operators. Nonetheless, idSTL efficiently finds a satisfying nominal trajectory (Figure \ref{fig:quad_demo}). It takes less than $15$s in MATLAB using the Z3 SMT solver \cite{de2008z3} and Gurobi LP solver \cite{gurobi}. All experiments were executed on an Intel Core i7 2.8-GHz processor with 32GB of memory.

\section{CONCLUSIONS}\label{sec:conclusion}

We present an efficient method for symbolic from STL specifications, suitable for complex and high-dimensional robotic applications. We propose a novel existential bounded model checking encoding for STL specifications with dynamic constraints, and harness highly efficient optimization methods along with iterative-deepening search to outperform state-of-the-art STL symbolic control methods. Furthermore, we prove the soundness and completeness of our algorithm and demonstrate its effectiveness in a simulation.

\bibliographystyle{IEEEtran}
\bibliography{IEEEabrv,Mendeley}

\printnoidxglossary[sort=word]
\printnoidxglossary[type=symbols,sort=word]

\end{document}